\documentclass[]{bytedance}
\usepackage{amsmath}
\usepackage{amssymb}
\usepackage{amsfonts}
\usepackage{amsmath}
\usepackage{microtype}
\usepackage{hyperref}
\usepackage{url}
\usepackage{booktabs}
\usepackage{cleveref}
\usepackage{graphicx}
\usepackage{multirow}
\usepackage{makecell}
\usepackage{array}
\usepackage[table]{xcolor}
\usepackage{colortbl}

\usepackage{lineno}

\definecolor{darkblue}{rgb}{0, 0, 0.5}
\hypersetup{colorlinks=true, citecolor=darkblue, linkcolor=darkblue, urlcolor=darkblue}

\title{PerceptionDLM: Parallel Region Perception with \\ Multimodal Diffusion Language Models}

\author{
\centerline{
Peking University MSALab, ByteDance
}
}

\abstract{
Multimodal large language models (MLLMs) have achieved remarkable progress in visual understanding tasks. 
However, most existing MLLMs rely on autoregressive generation, which limits their efficiency for perception tasks that require captioning multiple regions. 
In this work, we propose \textbf{PerceptionDLM}, a multimodal diffusion language model optimized for efficient parallel region perception. Built upon \textbf{PerceptionDLM-Base}, a strong foundational baseline that achieves state-of-the-art performance among open-source diffusion MLLMs, our architecture fully leverages the parallel decoding nature of DLMs. Specifically, we introduce efficient prompting and structured attention masking to enable simultaneous perception of multiple masked regions, allowing the model to generate region descriptions in parallel at both the sequence and token levels.
This design significantly improves inference efficiency compared with existing approaches that process regions sequentially.
To systematically evaluate the parallelism property of visual perception capability for DLMs, we construct a new \textbf{Para}llel \textbf{D}etailed \textbf{L}ocalized \textbf{C}aptioning Benchmark (\textbf{ParaDLC-Bench}) by scaling the DLC-Bench to include multiple region masks per image, enabling joint evaluation of both caption quality and inference efficiency. 
Experiments demonstrate that PerceptionDLM maintains competitive performance in region captioning while achieving substantial speed improvements for multi-region perception tasks. 
Our results highlight the potential of multimodal diffusion language models for efficient, parallel visual perception.
To the best of our knowledge, we are the first to achieve parallel region caption and perception by leveraging the advantages of diffusion language models.
Code, models, and datasets are released.
}

\date{\today} \correspondence{\email{yhtong@pku.edu.cn, sunyang2533@gmail.com}}
\checkdata[code]{\url{https://github.com/MSALab-PKU/PerceptionDLM}}
\checkdata[models]{\url{https://huggingface.co/collections/MSALab/perceptiondlm-model-zoo}}

\begin{document}

\maketitle

\section{Introduction}
\label{sec:intro}

\begin{figure*}[t!]
    \centering
    
    % --- Top Image (a) ---
    \begin{minipage}{1\linewidth}
        \centering
        \includegraphics[width=1.\linewidth]{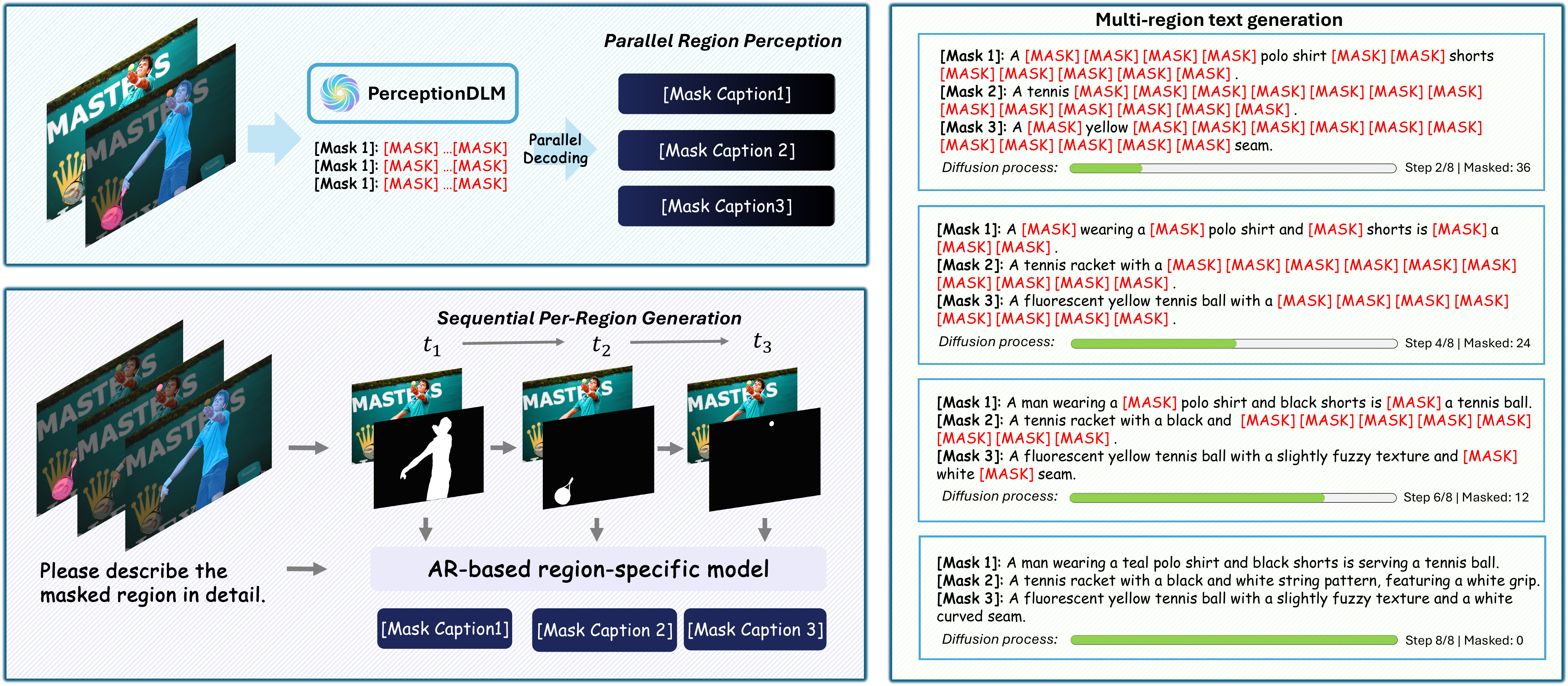}
        \vspace{1mm}
        \centerline{(a) Parallel Region Captioning with PerceptionDLM}
    \end{minipage}

    \vspace{4mm}

    % --- Bottom Left Image (b) ---
    \begin{minipage}{0.48\linewidth}
        \centering
        \includegraphics[width=\linewidth]{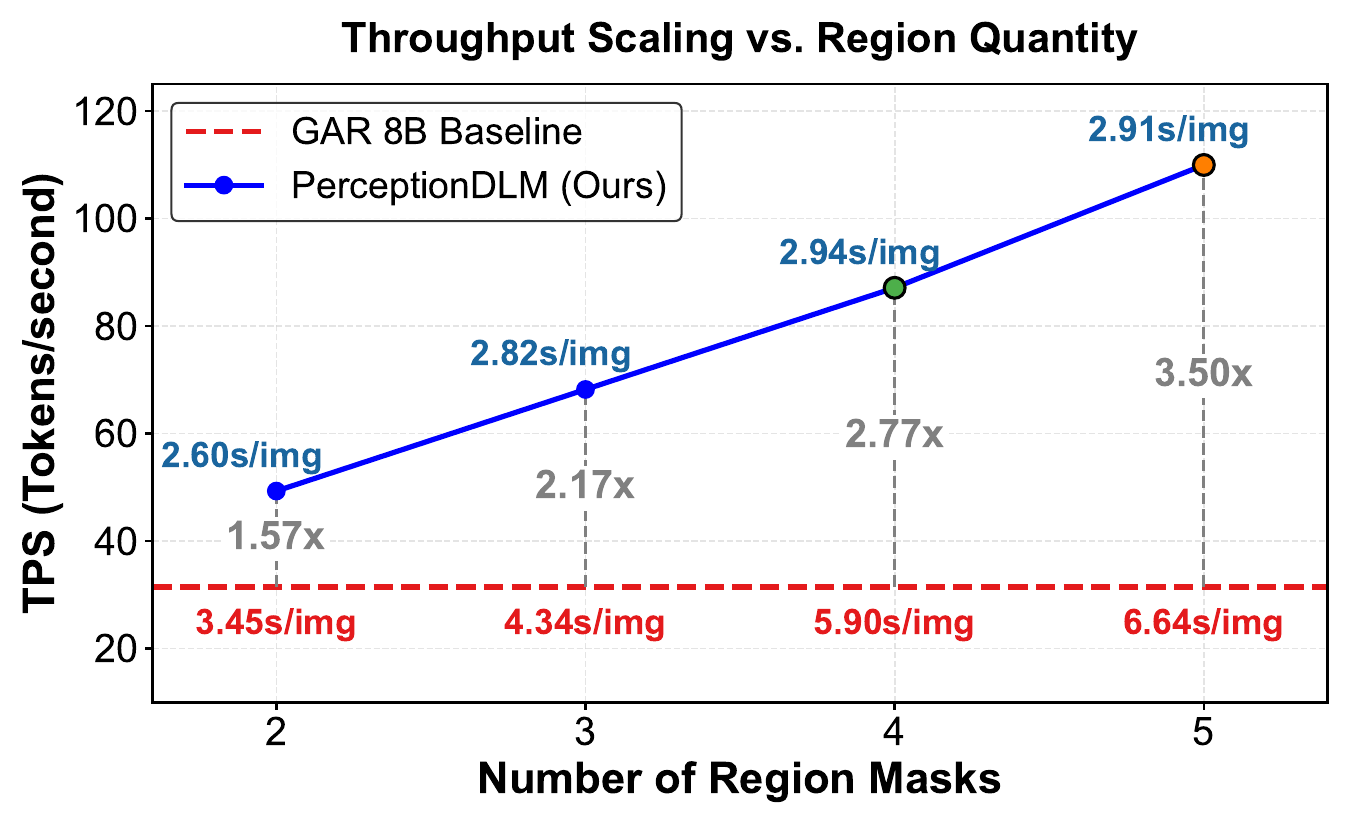}
        \vspace{1mm}
        \centerline{(b) Throughput vs. Region Quantity}
    \end{minipage}
    \hfill 
    % --- Bottom Right Image (c) ---
    \begin{minipage}{0.48\linewidth}
        \centering
        \includegraphics[width=\linewidth]{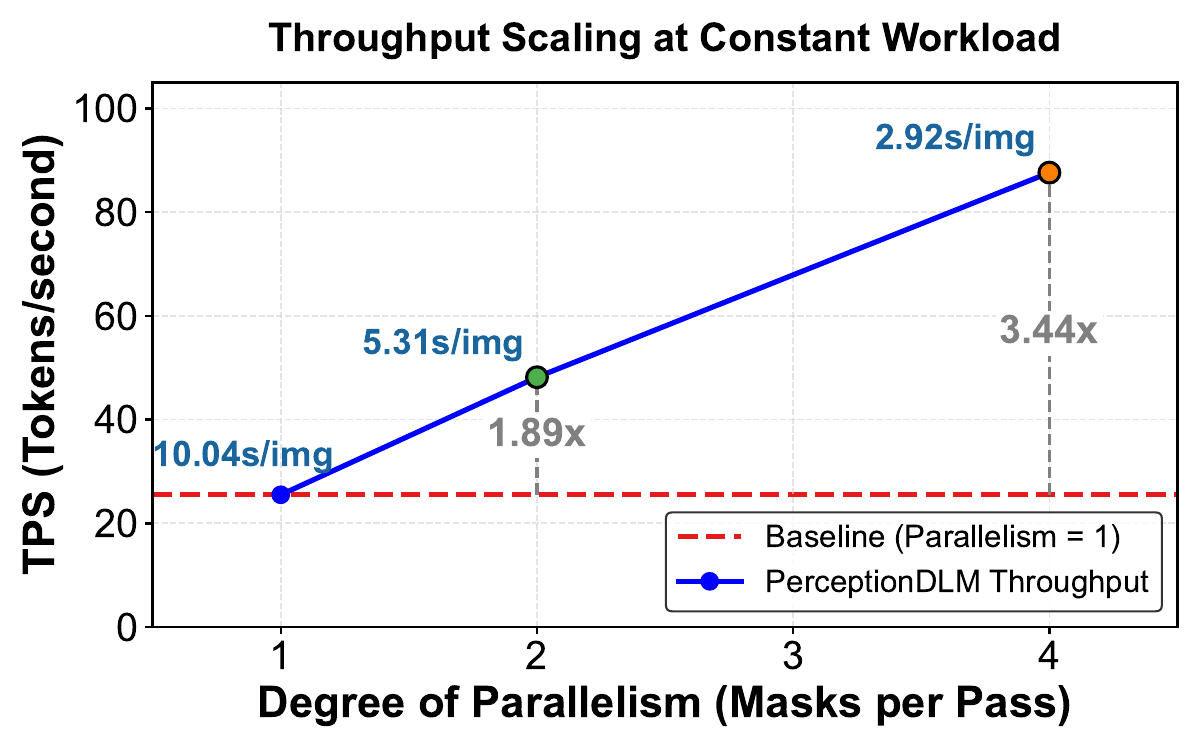}
        \vspace{1mm}
        \centerline{(c) Throughput Scaling at Constant Workload}
    \end{minipage}
    
    \vspace{2mm}
    
    \caption{\textbf{Overview and efficiency analysis of PerceptionDLM.} 
    \textbf{(a)} Given an image and multiple region masks, PerceptionDLM generates captions for all regions in parallel through a denoising diffusion process. In contrast to conventional AR-based models like DAM~\citep{lian2025describe} and GAR~\citep{wang2025grasp} that process regions sequentially, our approach entirely avoids the linear growth in inference cost. 
    \textbf{(b)} Throughput (TPS) scaling comparison. PerceptionDLM achieves near-linear TPS growth with stable per-image latency as the number of region masks increases. 
    \textbf{(c)} Parallelism scaling under a constant workload (4 masks per image), demonstrating up to a $3.44\times$ throughput speedup when fully parallelized.}
    \label{fig:teaser}
\end{figure*}

% \begin{figure*}[t]
%     \centering
%     \includegraphics[width=1.\linewidth]{figs/teaser2.pdf}
%     \caption{Illustration of \textbf{PerceptionDLM}, our parallel region captioning model built upon \textbf{PerceptionDLM-Base}. Given an image and multiple region masks, PerceptionDLM generates captions for all regions in parallel through a denoising diffusion process with token-level and sequence-level parallelism. In contrast, conventional AR-based region captioning models, such as DAM~\citep{lian2025describe} and GAR~\citep{wang2025grasp}, process regions sequentially, resulting in linear growth in inference cost as the number of regions increases.}

%     %\vspace{-2mm}
    
%     \label{fig:teaser}
% \end{figure*}

Visual perception~\citep{lin2014microsoft} is increasingly central to real-world multimodal intelligence. 
As applications demand more sophisticated interactions with the visual world, MLLMs have recently achieved remarkable progress across a wide range of visual understanding tasks via visual instruction-tuning and reinforcement learning pipelines~\citep{liu2023visual, li2025llava, an2025llava, bai2025qwen3, wang2025ross3d, zhu2025internvl3, lei2025scalability, meng2025open}.
However, while these models excel at image-level comprehension, many perception-heavy applications demand fine-grained, localized understanding. 
In these scenarios, a model must accurately describe multiple specific regions within a single image with fine-grained attributes and minimal confusion across targets~\citep{rasheed2024glamm, lian2025describe, wang2025grasp}. 
Consequently, fine-grained region-level captioning has emerged as a core capability for next-generation multimodal systems.
%AR decoding is inherently sequential at the token level. As a result, latency scales approximately linearly with the number of RoIs, resulting a bottleneck for dense multi-region perception.

Despite high accuracy, existing MLLMs are still dominated by autoregressive (AR) decoding, which introduces an efficiency bottleneck for multi-region perception. 
Under the AR paradigm, descriptions are typically generated sequentially for each individual region, and each description is token-by-token~\citep{lian2025describe, wang2025grasp}. 
As the number of queried regions increases, inference cost and latency grow rapidly, making the dense-region perception difficult to scale.

Recently, large diffusion language models (DLMs) have emerged as a competitive alternative to AR language modeling~\citep{nielarge, you2025llada, ye2025dream, yangmmada}. Their masked denoising generation paradigm enables non-autoregressive generation and exposes intrinsic token-level parallelism. 
Yet directly extending diffusion-based VLMs to fine-grained localized perception remains non-trivial: existing multimodal diffusion language models often lack strong perceptual capabilities, and their potential for concurrent multi-region perception remains underexplored~\citep{you2025llada, cheng2025sdar, ye2025dream, yangmmada}. 
This motivates the question of this work: \textit{Can we design a multimodal diffusion language model that preserves strong perception quality while unlocking practical parallelism for region-conditioned captioning?}

To answer this question, we propose \textbf{PerceptionDLM}, a diffusion-based framework for parallel region perception. 
We first build \textbf{PerceptionDLM-Base}, a strong discrete diffusion multimodal baseline that integrates a pretrained vision encoder with a diffusion language backbone via visual instruction tuning. 
Building upon this foundation, we introduce region-aware structural design. As illustrated in~\Cref{fig:teaser}, this enables PerceptionDLM to generate multiple region descriptions jointly within a single denoising process rather than through independent per-region decoding passes.

To evaluate both quality and efficiency under this setting, we introduce \textbf{ParaDLC-Bench}, which extends DLC-Bench~\citep{lian2025describe} from single-region evaluation to concurrent multi-region evaluation. 
ParaDLC-Bench explicitly measures caption quality and inference efficiency, enabling direct comparison between sequential and parallel decoding paradigms. 
%

%Extensive experiments show that PerceptionDLM establishes a new state-of-the-art among open discrete diffusion VLMs, achieving an average accuracy of 62.4\% on ParaDLC-Bench, which nearly doubles the performance of existing baselines like LLaDA-V (35.2\%). 
Extensive experiments show that PerceptionDLM provides a strong baseline among open discrete diffusion VLMs. On our 16-benchmark multimodal evaluation, PerceptionDLM-Base outperforms LLaDA-V~\citep{you2025llada} on 15 of 16 benchmarks. On ParaDLC-Bench, PerceptionDLM achieves an average accuracy of 62.4\%, which nearly doubles the performance of existing baselines like LLaDA-V (35.2\%).
Furthermore, it achieves competitive accuracy against strong AR region-specific models while substantially improving inference efficiency. As the number of queried regions increases, PerceptionDLM avoids the linear latency growth of AR models, yielding up to a 3.5 times throughput speedup in dense perception scenarios (e.g., 5 masks per image) under comparable model scales.

% Extensive experiments show that PerceptionDLM establishes a strong state-of-the-art baseline among open discrete diffusion-based MLLMs and achieves competitive accuracy against strong AR region-specific models, while substantially improving inference efficiency under comparable model scales.
%
Our main contributions are summarized as follows:
\begin{itemize}
    \item We present \textbf{PerceptionDLM-Base}, a high-performance, open discrete diffusion multimodal baseline that enhances perception capabilities for diffusion VLMs.
    \item We propose a \textbf{parallel region perception} mechanism for DLMs that combines region-aware mask embeddings and structured attention masking to enable simultaneous multi-region caption generation.
    \item We build \textbf{ParaDLC-Bench}, a benchmark tailored for multi-region localized captioning that evaluates both quality and efficiency.
    \item We demonstrate that \textbf{PerceptionDLM} achieves a strong trade-off between accuracy and efficiency, providing competitive region-caption quality with effective inference-speed gains over sequential region-captioning pipelines.
\end{itemize}
% \section{Method}
% \label{sec:method}
\vspace{-3mm}
\section{PerceptionDLM-Base: Stronger DVLM Baseline}

To establish a strong and scalable baseline for diffusion-based multimodal perception, we first introduce PerceptionDLM-Base, a multimodal diffusion language model that extends large-language diffusion models to visual instruction tuning. 

% \lxt{You can merge both preliminary and architecture into one. You can first introduce DLLM with notations and introduce what we have added on.}
\subsection{Diffusion Language Modeling and Multimodal Architecture}

\noindent
\textbf{Diffusion Language Modeling Preliminaries}
Let $x_0 = (x_1, \dots, x_N)$ be a sequence of discrete text tokens from a vocabulary $\mathcal{V}$. Discrete Diffusion Language Models (DLMs), such as LLaDA~\citep{nielarge}, formulate text generation as a generative Markov process. During the forward process, clean tokens $x_0$ are progressively corrupted into a sequence $x_t$ at timestep $t \in (0, 1]$, typically by replacing tokens with a special absorbing state (e.g., a \texttt{[MASK]} token). The reverse process learns a neural network $p_\theta(x_0 | x_t)$ to denoise $x_t$ back to $x_0$. The model is optimized using a reweighted variational lower bound, which simplifies to predicting the masked tokens:
\begin{equation}
\mathcal{L}_{\mathrm{DLM}} = -\mathbb{E}_{t, x_0, x_t} \left[ \frac{1}{t} \sum_{i=1}^{N} \mathbf{1}[x_t^i = \texttt{[MASK]}] \log p_\theta(x_0^i|x_t) \right],
\end{equation}
where $N$ is the sequence length and the indicator function $\mathbf{1}[\cdot]$ ensures that the loss is computed only for masked tokens.

\noindent
\textbf{Multimodal Architecture and Visual Instruction Tuning.}
% PerceptionDLM builds upon a diffusion language backbone (LLaDA-8B) and follows a widely adopted LLaVA-style design in multimodal large language models~\citep{liu2023visual}, which consists of three components: a pretrained vision encoder, a lightweight vision-language connector, and a LLM decoder.
To extend this diffusion language modeling paradigm to multimodal perception via visual instruction tuning~\citep{liu2023visual}, we introduce visual conditioning into the diffusion process. The overall architecture follows a widely adopted design comprising three components: a pretrained vision encoder, a lightweight vision-language connector, and a DLM decoder (LLaDA-8B).

Given an input image $X_v$ and a corresponding textual instruction $X_q$, the objective is to generate the target text response $X_a$. First, we utilize a pretrained SigLIP-2~\citep{tschannen2025siglip} as the vision encoder to extract visual features $Z_v = \Phi_v(X_v)$. A two-layer MLP with GELU activation serves as the connector $\Phi_{c}$, projecting the visual features into the LLM's text embedding space to obtain continuous visual embeddings $H_v = \Phi_c(Z_v)$.

The visual embeddings $H_v$ are concatenated with the discrete text embeddings of the instruction $X_q$ and the response $X_a$ to form the complete input sequence. During training, we apply the diffusion forward process \emph{only} to the target response tokens $X_a$, leaving the image representations $H_v$ and instruction tokens $X_q$ uncorrupted as conditions. Our multimodal visual instruction tuning objective is thus formulated as:
\begin{equation}
    \label{eq2}
    \mathcal{L}_{\mathrm{PerceptionDLM_{Base}}} = - \mathbb{E}_{(X_v, X_q, X_a), t, x_t} \left[ \frac{1}{t} \sum_{i \in \mathcal{M}_a} \log p_\theta(x_0^i | x_t, H_v, X_q) \right],
\end{equation}
where $x_t$ is the partially masked target response sequence, and $\mathcal{M}_a$ denotes the indices of the masked tokens strictly within $X_a$. 

PerceptionDLM-Base is trained using this discrete diffusion loss, followed by a 4-stage training paradigm. The detailed training parameters of each stage are shown in~\Cref{tab:traing_params}.

\noindent
\textbf{Dynamic Resolution for Multimodal Data}
To better handle high-resolution images and preserve fine-grained spatial details, we adopt a dynamic-resolution strategy following prior work~\citep{chen2024expanding}. Specifically, each input image is dynamically partitioned into a grid of tiles of $512\times512$ pixels based on the aspect ratio and resolution of the raw images. Optionally, if the number of tiles $n_{tiles} > 1$, we append an additional thumbnail of the original image to the end of the tiles list.  Then each tile is independently processed with pixel unshuffle operation to reduce the number of visual tokens to one-quarter of the original. 
Then, the preprocessed visual tokens are encoded by the vision encoder and concatenated into a sequence. 
This design allows the model to scale to arbitrary image resolutions while maintaining a balance between computational efficiency and spatial fidelity. 

%
% For the vision encoder and vision-language connector, we selected Siglip2(\citet{tschannen2025siglip}) and a two-layer MLP with GELU activation, respectively. 
%

%
% Our model is trained using the same discrete diffusion loss as LLaDA, followed by a 4-stage training paradigm. The detailed training parameters of each stage are shown in \Cref{tab:traing_params}.

% figure 1(option)
\vspace{-2mm}
\subsection{Training Strategies}

% \subsubsection{PerceptionDLM Training Pipeline}

We adopt a multi-stage training strategy for PerceptionDLM-Base, following the visual instruction tuning paradigm adapted for diffusion language models~\citep{you2025llada, liu2023visual}. 
Our training pipeline progressively improves multimodal alignment, general knowledge, and instruction-following ability in four stages.
We use the latest open-source datasets to conduct our training.
% Follwing LLaVA-OneVision-1.5 \citet{an2025llavaonevision15fullyopenframework} and Bee \citet{zhang2026beehighqualitycorpusfullstack}, we collect 52M open-source multimodal data to train

\noindent
\textbf{Stage 1: Vision-Language Alignment.}
We first perform a lightweight alignment between visual and textual representations using the Bee-Training-Data-Stage1~\citep{zhang2025bee}. At this stage, we primarily train the vision-language connector while keeping most of the backbone parameters frozen, enabling stable alignment between visual features and the language embedding space.

\noindent
\textbf{Stage 2: Middle-stage Training.}
After alignment, we conduct large-scale training on Bee-Training-Data-Stage2~\citep{zhang2025bee}. 
We explore two training strategies at this stage and the subsequent stages: (1) full-parameter training, where both the diffusion backbone and vision encoder are updated, and (2) partially frozen training, where the vision encoder remains fixed while updating the language model and projection layers. More details can be found in Appendix~\ref{sec:more_exp}.
This stage serves as a large-scale knowledge injection phase, exposing the model to diverse visual concepts and improving its general multimodal understanding capability. 
As in prior works~\citep{li2025llava, zhang2025bee}, such large-scale mid-training helps bridge the gap between lightweight alignment and instruction tuning by enhancing representational capacity.

\noindent
\textbf{Stage 3: Instruction Tuning.}
We then perform supervised fine-tuning (SFT) using 22M samples from the LLaVA-OneVision-1.5-Instruct-Data~\citep{an2025llava}, which contains diverse visual instruction tasks, including visual question answering, reasoning, OCR, and grounding. 
This stage enables the model to follow complex multimodal instructions and improves its generalization across a wide range of downstream tasks.

\noindent
\textbf{Stage 4: High-Quality SFT Refinement.}
Finally, we further refine the model using Honey-Data-15M~\citep{zhang2025bee}, a high-quality multimodal instruction dataset enriched with dual-level chain-of-thought annotations. 
This dataset emphasizes complex reasoning and long-form responses, which significantly enhance the model's reasoning ability.
%
% and instruction-following performance. 
%
This final stage improves both response quality and robustness in challenging multimodal scenarios.
%, as shown in ~\Cref{sec:exp}.

% Formulation.
% Dataset/Benchmark.
% Our Methods.
% Training and Inference.

\begin{figure*}[t]
    \centering
    \includegraphics[width=1.\linewidth]{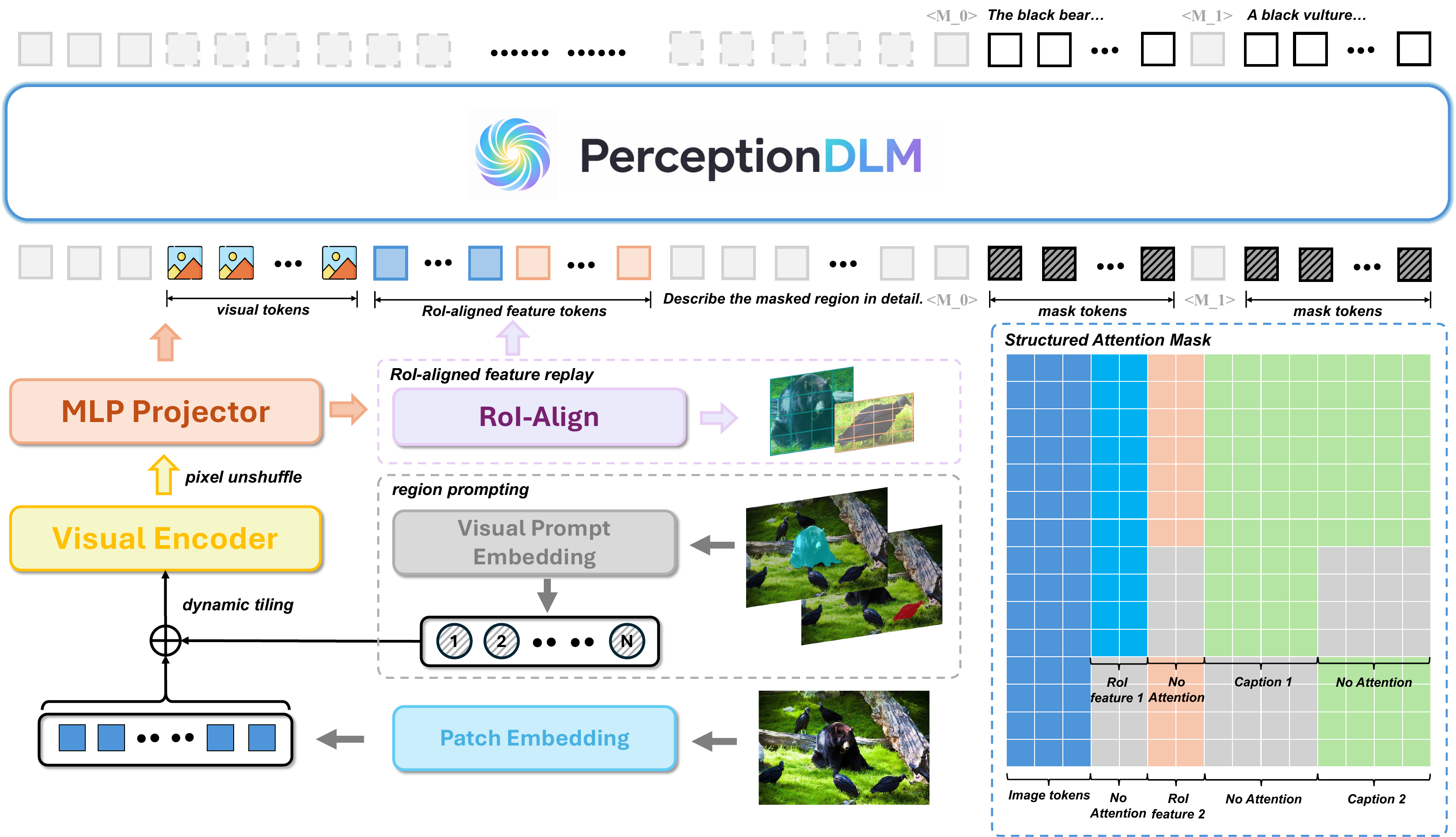}
    \caption{\textbf{Overview of the proposed parallel region perception architecture.} Built upon PerceptionDLM-Base, the model integrates region prompting, RoI-aligned feature replay, and a structured attention masking mechanism to enable parallel, disentangled caption generation for multiple regions within a single diffusion process.}
    %\caption{\textbf{Overview of the proposed parallel region perception architecture built upon PerceptionDLM.} Given an input image and multiple region masks, the visual encoder extracts global visual tokens, while region-specific masks are converted into mask embeddings and RoI-aligned feature tokens via region prompting and feature replay. Each region is associated with a dedicated placeholder token, which is expanded into localized RoI feature tokens that provide fine-grained visual context. During generation, we allocate separate masked token spans for each region caption. A structured attention masking mechanism is applied so that tokens in each region can attend only to shared global context and their own RoI features, while being isolated from other regions. This results in a block-wise attention structure that enables parallel and disentangled caption generation for multiple regions within a single diffusion process.}

    %\vspace{-2mm}
    
    \label{fig:arch}
\end{figure*}

\vspace{-3mm}
\section{Multimodal Diffusion Language Model for Parallel Perception}

\vspace{-2mm}
\subsection{Task Formulation}
\noindent
\textbf{Motivation.}
While recent MLLMs have achieved strong performance on visual understanding tasks, they typically rely on autoregressive decoding and generate outputs sequentially. 
This limitation becomes particularly evident in region-level perception tasks, where \textbf{multiple} regions need to be perceived and described.
In contrast, diffusion language models naturally support parallel token generation, enabling joint prediction of multiple outputs within a \textbf{single denoising process}, whereas recent methods~\citep{lian2025describe, wang2025grasp} can only process each region independently. 
This motivates us to revisit region-level perception from the perspective of parallel generation.

\noindent
\textbf{Problem Definition.}
Given an image $I$ and a set of $N$ regions $\{ R_i \}_{i=1}^N$, each represented by a binary mask, the goal is to generate a corresponding textual description $\{ y_i \}_{i=1}^N$ for all regions. 
Existing approaches typically process each region independently: $y_i = f(I,R_i)$, which leads to linear growth in inference latency with respect to the number of regions~\citep{rasheed2024glamm,lian2025describe}. 
In contrast, leveraging the inherent parallel decoding ability of diffusion language models, we can model $\{ y_i \}_{i=1}^N = f(I, \{ R_i \}_{i=1}^N)$, enabling simultaneous captioning of all region descriptions within a single diffusion process.

\vspace{-2mm}
\subsection{Model Architecture}
To enable efficient parallel region perception and captioning under the formulation above, we build upon the region-centric feature extraction paradigm of a recent strong AR-based baseline~\citep{wang2025grasp}. On top of this foundation, we propose a unified architecture that introduces region prompting and structured attention masking, as illustrated in~\Cref{fig:arch}.
Note that we highlight each component in a different color within the dashed boxes.

\noindent
\textbf{RoI-aligned Feature Replay from AR Baseline.} 
% To accurately represent region-specific visual information, we adopt the RoI-aligned feature extraction strategy introduced by the strong AR-based baseline, GAR~\citep{wang2025grasp}.
%
For each region mask, localized visual features are extracted directly from the vision encoder and projected into the language embedding space as placeholder tokens. During preprocessing, each placeholder is expanded into a set of RoI feature tokens extracted from the corresponding visual region.

\noindent
\textbf{Region Prompting.}
While RoI features capture visual details, parallel generation requires the model to strictly distinguish multiple concurrent targets. To effectively encode region identity into the input, we design a continuous region prompting mechanism. 
For each region $R_i$, we associate a learnable embedding $e_i$, which serves as a region-specific visual prompt. 
These embeddings are spatially broadcast and fused directly with the visual tokens corresponding to the masked regions, allowing the model to identify different regions and guide the generation process.
%By explicitly injecting these identity embeddings, the model can unambiguously differentiate between multiple targets, effectively grounding the parallel text generation process and preventing semantic mismatch across different regions.

% \noindent
% \textbf{RoI-aligned Feature Replay.}
% To represent region-specific visual information, we adopt a region-centric feature extraction strategy inspired by prior work~\citep{wang2025grasp}. 
% %
% For each region mask, we extract localized visual features from the vision encoder and project them into the language embedding space. 
% %
% This design allows the model to access both global image context and localized region features simultaneously. 
% %
% Concretely, we extend the input prompt by inserting special placeholder tokens, each corresponding to a specific region. 
% %
% During preprocessing, each placeholder is expanded into a set of RoI feature tokens extracted from the corresponding region.

\noindent
\textbf{Structured Attention Masking Mechanism.}
Since DLMs denoise all tokens for all regions simultaneously, a key challenge in parallel generation is preventing interference across regions. 
To address this problem, we design a structured attention masking mechanism that enforces region-wise independence while preserving global context. Specifically, for tokens corresponding to region $R_i$, we restrict their attention to: (1) global visual tokens, (2) shared textual prompt tokens, (3) RoI feature tokens associated with region $R_i$, and (4) tokens within the same region-specific caption span. Meanwhile, attention to RoI features and caption tokens of other regions is masked out. This structured design results in a block-wise attention pattern that enforces region-level independence while preserving shared global context, enabling accurate parallel caption generation.

As illustrated in~\Cref{fig:teaser}, our model can simultaneously describe multiple masked regions while achieving substantially better perception efficiency. Overall, our architecture transforms region-level perception from a sequential process into a structured parallel generation problem, fully leveraging the intrinsic parallelism of diffusion language models.

%\lxt{Missing training and inference details.}
%\lxt{This part is extremely important since this is the key difference of our approaches.}
%\lxt{In addition, there are no discussions on the generated token speed.}

% teaser

%
\vspace{-2mm}
\subsection{ParaDLC-Bench}
\vspace{-2mm}
Inspired by DLC-Bench~\citep{lian2025describe}, we introduce an extended evaluation benchmark, \textbf{ParaDLC-Bench}, specifically adapted for multi-mask scenarios. Traditional evaluations rely on comprehensive reference captions, which struggle to handle the complex semantic entanglement that occurs during concurrent multi-target generation. 
Our benchmark inherits the reference-free evaluation paradigm of DLC-Bench, utilizing an LLM as a judge, but expands upon it: we extend the core focus of the evaluation from intra-region detail accuracy to inter-region feature independence and anti-interference capability.
Similar to DLC-Bench, our evaluation process consists of two steps:
\begin{enumerate}
    \item The model is prompted to generate detailed descriptions in parallel for multiple masked regions within a single image from the benchmark dataset.
    \item An LLM serves as a judge, assessing the generated descriptions based on a predefined set of positive and negative questions. 
    Furthermore, we observed that the reliability of the evaluation heavily depends on the reasoning capabilities of the judge LLM itself. Therefore, compared to DLC-Bench, which employs Llama-3.1-8B~\citep{grattafiori2024llama}, we utilize the more advanced GPT-5.2~\citep{gpt5.2} as our judge model to ensure accurate and robust assessments in complex multi-target interaction scenarios.
\end{enumerate}

Regarding question design, we adopt the strategy from DLC-Bench and tailor it for parallel caption generation tasks:
\begin{itemize}
    \item \textbf{Positive questions:} Focus on specific attributes unique to the target mask. The model earns a point for accurate inclusion, receives no points for omission, and incurs a penalty for factual errors.
    \item \textbf{Negative \& Interference questions:} We introduce a key innovation within this category specifically for multi-target scenarios. In addition to checking for the inclusion of typical but absent attributes as in DLC-Bench, we specifically examine whether the model hallucinates features from other concurrent masked objects in the image into the current target's description (i.e., attribute entanglement). A point is awarded if the model successfully avoids such cross-region hallucination; otherwise, it incurs a penalty. To prevent completely off-topic captions from receiving high scores, this point is only awarded if the model correctly recognizes the target object.
\end{itemize}

To ensure the reliability of the evaluation, we implemented a rigorous quality control process. Expert human annotators conducted multiple rounds of strict cross-verification on all candidate questions to eliminate ambiguity, ensure strict spatial alignment with the targeted visual regions, and verify that the negative distractors were genuinely deceptive. Furthermore, we verified the robustness of our benchmark by re-evaluating the models across different open-source and closed-source judge LLMs (e.g., Qwen3.5-27B and Gemini-3.1-Pro), demonstrating stable performance rankings (detailed in Appendix~\ref{sec:more_exp}).
With this design, ParaDLC-Bench not only retains the flexibility and accuracy of DLC-Bench but also effectively tests the model's true competence in handling dense, multi-target tasks. 
Our benchmark ultimately comprises 2345 manually verified questions, covering a wide range of multi-region attribute interactions and potential cases of hallucination.
For more benchmark details, including the annotation pipeline and statistics, please refer to Appendix~\ref{sec:detail_bench}.

\vspace{-2mm}
\subsection{Training Data Engine}
% \vspace{-2mm}
To enhance parallel-region captioning of DLMs, we further construct high-quality single-image, multi-mask caption data, which we refer to as \textbf{ParaCaption-5.7M}.
We initially explore the Describe Anything dataset~\citep{lian2025describe}, which provides detailed caption annotations. 
However, because it lacks such concurrent multi-mask samples, we design an automated data construction pipeline based on existing large-scale segmentation datasets.

The training data comprises two different sources: the SA-1B dataset~\citep{kirillov2023segment} and the COCONut dataset~\citep{deng2024coconut}. For a selected subset of the massive SA-1B dataset, which contains numerous part-level regions unsuitable for general descriptions, we first filter out masks that are completely occluded (which are typically considered part-level). 
We then employ GAR-8B~\citep{wang2025grasp}, a state-of-the-art region captioning model, to generate initial descriptions and employ an LLM to extract the core categories. 
Next, using these extracted categories, we apply SAM3~\citep{carion2025sam} to re-predict the masks and discard low-matching samples by calculating the Intersection over Union (IoU) with the ground-truth masks. 
Meanwhile, for the COCONut dataset, which includes inherent mask and category annotations, we similarly use GAR-8B to generate initial descriptions and directly use Qwen3-8B~\citep{yang2025qwen3} to determine whether the generated text semantically matches the provided ground-truth category annotations.
All verified data pairs from both datasets undergo a unified post-processing stage for length restriction and anti-repetition (hallucination) filtering to yield the final caption texts. Ultimately, we obtained 334k images with 3.4M masks from the COCONut dataset, and 83k images with 2.3M masks from the SA-1B dataset. 
\vspace{-3mm}
\section{Experiment}
\label{sec:exp}

\begin{table*}[t!]
\centering
\caption{\textbf{Evaluation comparison across vision-language models on multimodal understanding benchmarks.} Most scores are reported as provided in the original notes. The symbol $^\dagger$ denotes results we evaluated using official1 checkpoints and inference scripts, while $^\star$ indicates results we evaluated from VLMEvalKit~\citep{duan2024vlmevalkit}. ``--'' indicates that the corresponding metric is not reported in the original paper.}
\label{tab:main_results}
\setlength{\tabcolsep}{4.5pt}
\renewcommand{\arraystretch}{1.25}
\resizebox{1\textwidth}{!}{
\begin{tabular}{l | >{\columncolor{blue!8}}c c c c c c | c c}
\toprule[1.5pt]
\multirow{2}{*}{\textbf{Benchmark}} & \textbf{PerceptionDLM-Base} & \textbf{LLaDA-V} & \textbf{MMaDA} & \textbf{LaViDa} & \textbf{SDAR-VL} & \textbf{Dream-VL} & \textbf{Qwen2.5-VL} & \textbf{InternVL3} \\
& 8B & 8B & 8B & 8B & 8B & 7B & 7B & 8B \\
\midrule
\multicolumn{9}{c}{\textbf{\textit{General VQA}}} \\
\midrule
MMStar & \textbf{63.7} & 60.1 & -- & -- & 59.9 & 59.9 & 63.9 & 68.2 \\
SeedBench & \textbf{78.9} & 74.8 & 64.2 & -- & 75.5 & 76.4 & 77.0$^\star$ & 77.1$^\star$ \\
MMBench & \textbf{85.0} & 82.9 & 68.5 & 73.8 & 82.2 & 83.0 & 83.5 & 83.4 \\
\midrule
\multicolumn{9}{c}{\textbf{\textit{Reasoning}}} \\
\midrule
MMMU & 47.2 & 48.6 & 48.6 & 30.2 & \textbf{53.0} & 52.2 & 51.3$^\star$ & 57.3$^\star$ \\
MathVista & \textbf{65.5} & 52.4$^\dagger$ & -- & 42.1 & 62.5 & 63.1 & 68.2 & 71.6 \\
MathVerse$_{\text{Vision\_Only}}$ & 25.3 & 20.6$^\dagger$ & -- & 27.2 & \textbf{36.5} & -- & 42.7$^\star$ & 23.1$^\star$ \\
\midrule
\multicolumn{9}{c}{\textbf{\textit{OCR \& Doc \& Chart}}} \\
\midrule
AI2D & \textbf{85.0} & 77.8 & -- & 69.0 & 79.9 & 81.2 & 83.9 & 85.2 \\
ChartQA & \textbf{91.6} & 78.3 & -- & 61.0 & 82.7 & 86.8 & 86.2 & 86.6 \\
DocVQA & 89.9 & 83.9 & -- & 56.1 & 88.3 & \textbf{94.4} & 94.9 & 92.7 \\
InfoVQA & 74.6 & 66.3 & -- & 36.2 & 73.2 & \textbf{81.4} & 82.6 & 76.8 \\
\midrule
\multicolumn{9}{c}{\textbf{\textit{Perception}}} \\
\midrule
MMVP & \textbf{82.0} & 76.7$^\star$ & -- & -- & 66.5 & -- & 73.3$^\star$ & 80.0$^\star$ \\
BLINK & \textbf{60.3} & 50.9 & -- & -- & -- & 52.9 & 55.3 & 55.5 \\
RealWorldQA & \textbf{73.7} & 63.2 & -- & -- & 66.5 & 66.3 & 68.4 & 70.8 \\
CV-Bench-2D & \textbf{79.8} & 77.7$^\star$ & -- & -- & -- & -- & 75.6$^\star$ & 79.0$^\star$ \\
\midrule
\multicolumn{9}{c}{\textbf{\textit{Others}}} \\
\midrule
HallusionBench & \textbf{58.4} & 50.9$^\star$ & -- & -- & 44.4 & -- & 51.9 & 49.9 \\
V$^\star$ & \textbf{73.3} & 62.8$^\star$ & -- & -- & -- & -- & 76.4$^\star$ & 67.5$^\star$ \\
\bottomrule[1.5pt]
\end{tabular}
}
\end{table*}

% As shown in \Cref{tab:main_results},... (compare main results with other DLMs).
% tabel2: main results on general 
\vspace{-2mm}
\subsection{Implementation Details of Parallel Region Perception Model}
\vspace{-2mm}
We adopt PerceptionDLM-Base as our base model for training a parallel region perception model, as it demonstrates strong perception capabilities across several frontier diffusion-based VLMs.
%During the training phase of PerceptionDLM, we set the maximum number of region prompts per image to 6 for training efficiency while ensuring the model learns to handle multi-target perceptual tasks effectively.
During training, we set the number of region prompts per image to 6 for training efficiency while ensuring the model learns multi-target perception. 
%We use six learnable region-ID embeddings $\{e_i\}_{i=1}^{6}$.
%
During evaluation, PerceptionDLM is under the default inference setting of 32 generation steps for a sequence length of 32 per mask. We use the same training loss as in~\Cref{eq2} with PerceptionDLM while setting all parameters trainable. We use transformers trainer and AdamW optimizer with a global batch size of 256 and a learning rate of $4 \times 10^{-5}$, with a linear warmup during the first $3\%$ steps followed by a cosine decay schedule for the remaining steps. For RoI-aligned feature replay, we set the RoI output size to $ 4 \times 4 $ in default. 

\begin{table*}[t!]
\centering
\setlength{\belowcaptionskip}{8pt}
\caption{\textbf{Comparison on the ParaDLC-Bench and DLC-Bench.}
The symbol $1^*$ indicates that during the inference of these baseline diffusion VLMs, we set the number of denoising steps equal to the generation length, aiming to achieve their best possible generation quality.
}
\label{tab:dlcbench_comparison}
\setlength{\tabcolsep}{8pt}
\renewcommand{\arraystretch}{1.15}
\resizebox{1\textwidth}{!}{
\begin{tabular}{l c ccc cc ccc}
\toprule[1.5pt]
\multirow{2}{*}{Method} & \multirow{2}{*}{Size} & \multicolumn{5}{c}{ParaDLC-Bench} & \multicolumn{3}{c}{DLC-Bench} \\
\cmidrule(lr){3-7} \cmidrule(lr){8-10}
& & Pos (\%) & Neg (\%) & Avg (\%) & TPF & Time (s) & Pos (\%) & Neg (\%) & Avg (\%) \\
\midrule
\multicolumn{10}{c}{\textbf{\textit{General VLMs}}} \\
\midrule
GPT-5.2 & -- & 38.0 & 71.0 & 55.2 & -- & -- & 18.0 & 60.8 & 39.4\\
Gemini-2.5-Pro & -- & 39.7 & 73.3 & 57.5 & -- & -- &29.8 & 66.0 & 47.9\\
Gemini-3.1-Pro & -- & 43.6 & 81.1 & 63.7 & -- & -- & 26.2 & 65.8 & 46.0 \\
\midrule
\multicolumn{10}{c}{\textbf{\textit{AR-based Region-specific VLMs}}} \\
\midrule
PixelRefer & 7B & 40.8 & 78.7 & 60.5 & 1 & 718 & 49.6 & 87.0 & 68.3 \\
DAM & 3B & 48.1 & 87.2 & 69.2 & 1 & 326 & 52.3 & 82.2 & 67.3 \\
GAR & 8B & 49.0 & 87.6 & 69.5 & 1 & 479 & 50.9 & 84.6 & 67.8 \\
\midrule
\multicolumn{10}{c}{\textbf{\textit{Diffusion-based VLMs}}} \\
\midrule
LLaDA-V & 8B & 24.1 & 46.3 & 35.2 & 1$^{*}$ & 3241 & 10.0 & 39.2 & 24.6 \\
SDAR-VL & 8B & 30.2 & 28.8 & 31.3 & 1$^{*}$ & 945 & 9.9 & 47.6 & 28.8\\
Dream-VL & 7B & 29.7 & 28.6 & 30.4 & 1$^{*}$ & 446 & 9.8 & 39.6 & 24.7 \\
\midrule
\multicolumn{10}{c}{\textbf{\textit{Ours Parallel Caption Model}}} \\
\midrule
\rowcolor{blue!8}
\textbf{PerceptionDLM} & 8B & 42.3 & 82.4 & 62.4 & 2.9 & 276 & 33.4 & 72.8 & 53.1\\
\bottomrule[1.5pt]
\end{tabular}
}
\end{table*}

%\subsection{Main Results}
\vspace{-2mm}
\subsection{Evaluation on Multimodal Benchmarks for PerceptionDLM-Base}
\vspace{-2mm}
% We evaluate PerceptionDLM on a diverse set of multimodal benchmarks covering general VQA, reasoning, chart and document understanding, perception-oriented tasks, and hallucination robustness. Specifically, we report results on \textbf{General VQA} benchmarks (MMStar, SeedBench, MMBench), \textbf{Reasoning} benchmarks (MMMU, MathVista, MathVerse-Vision-Only), \textbf{OCR, Chart and Document Understanding} benchmarks (AI2D, ChartQA, DocVQA, InfoVQA), \textbf{Perception} benchmarks (MMVP, BLINK, RealWorldQA, CV-Bench-2D), and \textbf{Others} (HallusionBench, CaptionQA, V*). As shown in \Cref{tab:main_results}, PerceptionDLM achieves strong performance across nearly all benchmark groups, demonstrating the effectiveness of diffusion-based multimodal training for general visual understanding.
% We evaluate PerceptionDLM across 16 diverse multimodal benchmarks. As shown in \Cref{tab:main_results}, PerceptionDLM achieves strong performance across these diverse categories, demonstrating the effectiveness of diffusion-based multimodal training for comprehensive visual understanding. Additional comparisons with more multimodal large language models will be presented in the Appendix [XX].
As shown in~\Cref{tab:main_results}, we evaluate PerceptionDLM-Base across 16 diverse multimodal benchmarks. Additional information and experiments are shown in Appendix~\ref{sec:more_exp}.

\noindent
\textbf{PerceptionDLM-Base consistently outperforms prior diffusion VLMs.}
PerceptionDLM-Base establishes a new, strong baseline among multimodal diffusion language models. Compared with frontier diffusion VLMs, our model delivers consistent and substantial gains on most benchmarks. Most notably, PerceptionDLM-Base outperforms LLaDA-V~\citep{you2025llada} on 15 out of the 16 evaluated benchmarks. Furthermore, when compared to more recent strong baselines such as SDAR-VL-8B~\citep{cheng2025sdar} and Dream-VL-7B~\citep{ye2025dream}, PerceptionDLM maintains a clear, comprehensive advantage, particularly in general VQA, fine-grained visual perception, and hallucination robustness.
These gains indicate that the proposed training pipeline and multimodal architecture significantly enhance diffusion-based models' capabilities for general-purpose understanding and perception.

\noindent
\textbf{Competitive performance with advanced autoregressive VLMs.}
PerceptionDLM-Base is also competitive with recent autoregressive VLMs of similar size. Across the 16 benchmarks, PerceptionDLM-Base achieves superior or comparable scores on a majority of the tasks. 
It excels particularly in fine-grained visual perception, systematically outperforming both Qwen2.5-VL-7B~\citep{bai2025qwen25vl} and InternVL3-8B~\citep{zhu2025internvl3} in these specific areas. 
This suggests that PerceptionDLM-Base possesses a distinct advantage in region-sensitive and detail-oriented visual understanding. While a performance gap remains in complex, reasoning-heavy scenarios (e.g., MMMU~\citep{yue2024mmmu} and MathVista~\citep{lu2023mathvista}), we observe that arbitrary-order parallel decoding fundamentally limits the reasoning potential of diffusion language models~\citep{ni2026flexibility}. Thus, we adopt autoregressive-order decoding for PerceptionDLM-Base during these mathematical reasoning evaluations to better preserve reasoning traces. Inspired by recent advancements like DeepSeek-R1~\citep{guo2025deepseek}, this bottleneck highlights a clear avenue for future work: leveraging Reinforcement Learning (RL) to further unlock the reasoning potential of diffusion-based VLMs.

% We will explore applying RL in future works.
% Overall, the results in \Cref{tab:main_results} show that PerceptionDLM not only advances the state of the art among open-source diffusion-based VLMs, but also narrows the gap to leading autoregressive models.

\vspace{-2mm}
\subsection{Evaluation on Captioning Benchmarks with PerceptionDLM}
\vspace{-2mm}
To further assess fine-grained multimodal understanding, we evaluate PerceptionDLM on region captioning benchmarks, including our multi-region ParaDLC-Bench and the single-region DLC-Bench~\citep{lian2025describe}.

\noindent
\textbf{Superior region captioning capability among diffusion VLMs.}
As shown in~\Cref{tab:dlcbench_comparison}, PerceptionDLM exhibits leading advantages over existing diffusion-based VLMs.On ParaDLC-Bench, it achieves an average accuracy of 62.4\%, nearly doubling the performance of SDAR-VL~\citep{cheng2025sdar} (31.3\%) and LLaDA-V~\citep{you2025llada} (35.2\%). This substantial margin is consistent on DLC-Bench~\citep{lian2025describe} (51.9\% vs. 24.6\% for baselines). 

\noindent
\textbf{Competitive performance with unprecedented efficiency.}
To analyze efficiency, we adopt the Tokens Per Forward (TPF) metric~\citep{qian2026d3llm}, which quantifies the average number of generated tokens per forward pass. Compared with AR-based region-specific models (e.g., DAM and GAR), PerceptionDLM demonstrates highly competitive accuracy while unlocking massive speed advantages. Although its average accuracy on ParaDLC-Bench is slightly lower than AR-based models, PerceptionDLM drastically reduces the total inference time across the benchmark to just 276 seconds, compared to 479 seconds for GAR and 718 seconds for PixelRefer~\citep{yuan2025pixelrefer}. 

This efficiency is driven by our parallel decoding paradigm. While AR models and standard DVLM baselines are restricted to a TPF of 1, PerceptionDLM achieves a TPF of 2.9, generating multiple region captions simultaneously. Note that on DLC-Bench, where instances contain only a single mask, our parallel-processing advantage cannot be fully exploited. Nevertheless, PerceptionDLM still sets a new benchmark for diffusion-based VLMs across both settings.

% \begin{figure*}[t!]
%     \centering\small
%     \begin{minipage}{0.48\linewidth}
%         \centering
%         \includegraphics[width=\linewidth]{figs/tps_vs_masks.pdf}
%         \centerline{(a) Throughput vs. Region Quantity}
%     \end{minipage}
%     \hfill % 使用 \hfill 撑开两张图之间的间距
%     \begin{minipage}{0.48\linewidth}
%         \centering
%         \includegraphics[width=\linewidth]{figs/parallelism_scaling.pdf}
%         \centerline{(b) Throughput Scaling at Constant Workload}
%     \end{minipage}
%     \caption{\textbf{Efficiency and scalability analysis of PerceptionDLM.} 
%     \textbf{(a)} Throughput (TPS) scaling with an increasing number of region masks per image. \textbf{(b)} Parallelism scaling under a constant workload (4 masks per image).}
%     \label{fig:efficiency_analysis}
% \end{figure*}

To systematically analyze the computational advantages of PerceptionDLM, we conduct a speed and efficiency profiling.
% As shown in \Cref{fig:efficiency_analysis}, conventional autoregressive models, such as GAR 8B~\citep{wang2025grasp}, suffer from a sequential bottleneck when processing multiple visual regions, causing per-image latency to scale linearly. In contrast, PerceptionDLM breaks this dependency through non-autoregressive parallel diffusion generation.
As shown in~\Cref{fig:teaser}(b), PerceptionDLM achieves near-linear TPS growth while maintaining stable per-image latency ($\sim$2.9s).
In contrast, GAR-8B is bottlenecked at a nearly constant TPS, and its latency degrades approximately linearly with the number of regions. 
As shown in~\Cref{fig:teaser}(c), under a constant heavy workload (4 masks per image), increasing the degree of parallelism (masks processed per pass) yields strong scaling for PerceptionDLM: throughput improves by 3.44 times, and single-image latency drops from 10.04s to 2.92s when fully parallelized.
\vspace{-3mm}
\section{Conclusion}
\label{sec:conclusion}
\vspace{-3mm}
We present PerceptionDLM, a diffusion-based multimodal model for parallel region perception.
Built on a stronger diffusion VLM baseline, it generates multiple region captions in a single denoising step rather than decoding them one by one.
This design preserves competitive caption quality while substantially improving efficiency on multi-region perception tasks.
We also introduce ParaDLC-Bench to evaluate both caption accuracy and inference speed in parallel localized captioning.
Overall, our results show that diffusion-based multimodal models are a promising direction for efficient fine-grained visual understanding.
We hope our work can inspire the community for a new way of caption generation.

% \section*{Ethics Statement}
% % \vspace{-3mm}
% Our work does not collect new data on humans and animals.
% %
% We use open-source datasets to conduct the DVLM experiments.
% %
% Although these datasets include humans and animals, our research only describes their appearance and does not cover gender, citizenship, or identity information.
%Authors can add an optional ethics statement to the paper. 
%For papers that touch on ethical issues, this section will be evaluated as part of the review process. The ethics statement should come at the end of the paper. It does not count toward the page limit, but should not be more than 1 page. 

% \clearpage
\section{Contributions}
\label{sec:contribution}

\textbf{Authors} Yueyi Sun$^{1,2,*}$, Yuhao Wang$^{1,2,*}$, Jason Li$^{2}$, Ye Tian$^{1}$, Tao Zhang$^{3}$, Jacky Mai$^{2}$, Yihan Wang$^{1}$, Haochen Wang$^{4}$, Jinbin Bai$^{5}$, Ling Yang$^{1}$, Yunhai Tong$^{1,\dagger}$

\vspace{2mm}

\textbf{Affiliations} $^{1}$Peking University \quad $^{2}$ByteDance \quad $^{3}$WHU \quad $^{4}$CASIA \quad $^{5}$ NUS

\vspace{2mm}

$^{*}$ Work was done during their internship at ByteDance\\
$^{\dagger}$ Corresponding author

\clearpage

\bibliography{colm2026_conference}
\bibliographystyle{colm2026_conference}
\clearpage

\beginappendix
\tableofcontents
\clearpage
% Overview.
% 1. Data Pipeline and Data Statistics.
% 2. More experiment results
% 3. Related Works.
% 4. Limitations and future works (directions).

\section{Overview}\vspace{-5pt}

% This appendix provides supplementary materials, extended discussions, and additional experimental results to support the main paper.
Here is the table of contents of this appendix:

\begin{itemize}
    \item In \textbf{Appendix~\ref{sec:more_exp}}, we provide more implementation details along with additional experimental results. Detailed ablation studies analyzing the impact of each core component of PerceptionDLM can also be found in this section.
    \item In \textbf{Appendix~\ref{sec:detail_bench}}, we introduce the details of our proposed ParaDLC-Bench, including the specific annotation guidelines, quality control protocols, and overall dataset statistics.
    \item In \textbf{Appendix~\ref{sec:related_works}}, we provide an extended discussion of related works, covering broader literature on diffusion language models and region-level multimodal perception and captioning.
    \item In \textbf{Appendix~\ref{sec:qua}}, we provide extensive visualization and qualitative results, showcasing PerceptionDLM's performance on fine-grained image understanding and parallel region captioning tasks.
    \item In \textbf{Appendix~\ref{sec:limitation}}, we discuss the potential limitations of our current approach and provide an analysis of typical failure cases.
\end{itemize}

\section{Additional Experiments}
\label{sec:more_exp}

\noindent
\textbf{Implementation details.} We summarize the implementation details of PerceptionDLM-Base pretraining and the parallel captioning model in this section. 

The four-stage training setup of PerceptionDLM-Base is listed in~\Cref{tab:traing_params}; the entire pipeline is trained on $32\times$NVIDIA H100 (80GB) GPUs and takes approximately three weeks in total. We use the AdamW optimizer with cosine learning-rate schedule and $3\%$ linear warmup, BF16 mixed precision, and gradient checkpointing. The vision encoder is SigLIP-2~\citep{tschannen2025siglip} and the diffusion backbone is initialized from LLaDA-Instruct-8B~\citep{nielarge}; images are processed with the dynamic-resolution tiling strategy using a tile size of $512\times512$.

For the parallel captioning model, we initialize from PerceptionDLM-Base and set all parameters trainable. We use a global batch size of 256, learning rate of $4\times10^{-5}$, one training epoch, maximum sequence length of 4096, and RoI output size of $4\times4$. We set the maximum number of region prompts per image at 6 during training. Training on the full ParaCaption-5.7M corpus takes about 2 days on $32\times$H100 GPUs and since it starts from the strong base checkpoint, this stage is substantially cheaper than base pretraining.

For parallel-captioning ablations, unless otherwise specified, all runs use the same setting: training on DAM~\citep{lian2025describe} only, default PerceptionDLM as baseline architecture, and inference with 32 diffusion steps per diffusion process and generation length 32 per mask.
All latency and throughput evaluations are conducted on a single H100 GPU using BF16 precision, with a strictly matched batch size of 1 per image for all models. We keep all other hyper-parameters identical across ablation runs to ensure fair comparison.
For evaluation protocols, multimodal benchmarks are run with VLMEvalKit~\citep{duan2024vlmevalkit}. For ChartQA, DocVQA, and InfoVQA, we additionally adopt a stricter answer-matching judge based on a locally deployed Qwen3-8B~\citep{yang2025qwen3} to better reflect the true accuracy. For region-captioning benchmarks, the LLM judge (GPT-5.2 by default) is queried at \texttt{temperature=0} to remove sampling noise, and we further report robustness across different judge models in~\Cref{tab:judge_sensitivity}.

\noindent
\textbf{Evaluation on Multimodal Benchmarks} We evaluate PerceptionDLM-Base on a diverse set of multimodal benchmarks covering general VQA, reasoning, chart and document understanding, perception-oriented tasks, and hallucination robustness. Specifically, we report results on \textbf{General VQA} benchmarks (MMStar~\citep{chen2024mmstar}, SeedBench~\citep{li2023seedbench}, MMBench~\citep{liu2024mmbench}), \textbf{Reasoning} benchmarks (MMMU~\citep{yue2023mmmu}, MathVista~\citep{lu2023mathvista}, MathVerse-Vision-Only~\citep{zhang2024mathverse}), \textbf{OCR, Chart and Document Understanding} benchmarks (AI2D~\citep{kembhavi2016ai2d}, ChartQA~\citep{masry2022chartqa}, DocVQA~\citep{mathew2021docvqa}, InfoVQA~\citep{mathew2021infographicvqa}), \textbf{Perception} benchmarks (MMVP~\citep{tong2024mmvp}, BLINK~\citep{fu2024blink}, RealWorldQA~\citep{xai2024realworldqa} , CV-Bench-2D~\citep{tong2024cambrian}), and \textbf{Others} (HallusionBench~\citep{guan2023hallusionbench}, V*~\citep{wu2023vstar}).
% As shown in \Cref{tab:main_results}, PerceptionDLM achieves strong performance across nearly all benchmark groups, demonstrating the effectiveness of diffusion-based multimodal training for general visual understanding.

%\input{tables/training_params}
\begin{table}[htbp]
\centering
\setlength{\belowcaptionskip}{8pt}
\caption{Training parameters of PerceptionDLM-Base}
\label{tab:traing_params}
\resizebox{\textwidth}{!}{
\begin{tabular}{l|c|c|c|c}
\toprule[1.5pt]
 & Stage-1 & Stage-2 & Stage-3 & Stage-4 \\
\midrule
Dataset & Bee-Training-Data-Stage1 & Bee-Training-Data-Stage2 & LLaVA-OneVision-1.5-Instruct-Data & Honey-Data-15M \\
\#Samples & 1M & 14M & 22M & 15M \\
\midrule
Vision Tower & 
siglip2-so400m-patch16-512 & 
siglip2-so400m-patch16-512 & 
siglip2-so400m-patch16-512 & 
siglip2-so400m-patch16-512\\
LLM Backbone & LLaDA-Instruct-8B & LLaDA-Instruct-8B & LLaDA-Instruct-8B & LLaDA-Instruct-8B \\
% Trainable Model Parameters & Projector & \multicolumn{3}{|c}{Projector + LLM backbone} \\
Trainable Model Parameters & Projector & Projector + LLM backbone & Projector + LLM backbone & Projector + LLM backbone \\
\midrule
Batch Size & 512 & 512 & 512 & 512 \\
Model Max Length & 2048 & 4096 & 4096 & 4096 \\
Learning Rate & $ 1 \times 10^{-3} $ & $ 4 \times 10^{-5} $ & $ 4 \times 10^{-5} $ & $ 4 \times 10^{-5} $ \\
Epoch & 1 & 1 & 1 & 1 \\
\bottomrule[1.5pt]
\end{tabular}
}
\end{table}

% \noindent
% \textbf{Ablation Study and Analysis} We provide additional ablations to analyze data scaling, vision encoder training strategy, component contribution, and decoding configuration.

\begin{table*}[htbp]
\centering
\setlength{\belowcaptionskip}{8pt}
\caption{Multimodal benchmark comparison between PerceptionDLM-Base with trainable ViT variant and the frozen-ViT baseline. The compared benchmarks follow the same protocol as in~\Cref{tab:main_results}.}
\label{tab:multimodal_vit_train_vs_freeze}
\setlength{\tabcolsep}{3pt}
\renewcommand{\arraystretch}{1.15}
\resizebox{0.95\textwidth}{!}{
\begin{tabular}{l | c c c c c c c c}
\toprule
\textbf{Model Variant} & \textbf{MMStar} & \textbf{MMBench} & \textbf{MMMU} & \textbf{MathVista} & \textbf{AI2D} & \textbf{ChartQA} & \textbf{MMVP} & \textbf{HallusionBench} \\
\midrule
PerceptionDLM-Base (Train ViT) & 61.9 & 83.7 & \textbf{48.7} & 61.9 & 82.0 &  89.3 & 79.7 & 54.6 \\
\rowcolor{blue!8}
\textbf{PerceptionDLM-Base (Freeze ViT, baseline)} & \textbf{63.7} & \textbf{85.0} & 47.2 & \textbf{65.5} & \textbf{85.0} & \textbf{91.6} & \textbf{82.0} & \textbf{58.4} \\
\bottomrule
\end{tabular}
}
\end{table*}

\noindent
\textbf{Impact of ViT Training Strategy on PerceptionDLM-Base.}
We explore two optimization strategies for instruction tuning: (1) full-parameter training (updating both the diffusion backbone and ViT), and (2) partially frozen training (freezing ViT).~\Cref{tab:multimodal_vit_train_vs_freeze} evaluates their impact on multimodal capabilities. The frozen-ViT baseline significantly outperforms the full-parameter variant across most benchmarks. Thus, we adopt the partially frozen strategy to preserve broad visual understanding.

%Although the full-parameter variant shows a marginal gain on the reasoning-heavy MMMU (48.7 vs. 47.2), its pervasive degradation on other tasks indicates that unfreezing the ViT compromises generalized pre-trained representations. Thus, we adopt the partially frozen strategy to preserve broad visual understanding.

% scaling data

% \begin{table}[htbp]
% \centering
% \setlength{\belowcaptionskip}{8pt}
% \caption{Data scaling ablation for Parallel Captioning. Increasing the amount and diversity of training data consistently improves performance.}
% % \vspace{2mm}
% \label{tab:data_scaling_ablation}
% \setlength{\tabcolsep}{6pt}
% \renewcommand{\arraystretch}{1.15}
% \resizebox{0.75\textwidth}{!}{
% \begin{tabular}{l | l | c c c}
% \toprule
% \textbf{Method} & \textbf{Training Data} & \textbf{Pos(\%) $\uparrow$} & \textbf{Neg(\%) $\uparrow$} & \textbf{Avg(\%) $\uparrow$} \\
% \midrule
% % 前三行的第一列留空
% & dam & 31.8 & 67.5 & 50.0 \\
% & dam + sam & 32.2 & 70.6 & 52.2 \\
% & dam + coconut & 39.0 & 73.0 & 56.9 \\
% \rowcolor{blue!8}
% % 在最后一行使用白底并向上合并 4 行
% \cellcolor{white}\multirow{-4}{*}{\textbf{PerceptionDLM}}
% & \textbf{dam + sam + coconut} & \textbf{39.7} & \textbf{80.6} & \textbf{61.4} \\
% \bottomrule
% \end{tabular}
% }
% \end{table}

\begin{table}[htbp]
\centering
\setlength{\belowcaptionskip}{8pt}
\caption{Data scaling ablations for Parallel Captioning. Increasing the amount and diversity of training data consistently improves performance.}
% \vspace{2mm}
\label{tab:data_scaling_ablation}
\setlength{\tabcolsep}{6pt}
\renewcommand{\arraystretch}{1.15}
\resizebox{0.7\textwidth}{!}{
\begin{tabular}{l | l | c c c}
\toprule
\textbf{Method} & \textbf{Training Data} & \textbf{Pos(\%) $\uparrow$} & \textbf{Neg(\%) $\uparrow$} & \textbf{Avg(\%) $\uparrow$} \\
\midrule
& DAM & 33.8 & 73.6 & 53.7 \\
& DAM + COCONut & 38.4 & 77.1 & 57.7 \\
\rowcolor{blue!8}
\cellcolor{white}\multirow{-3}{*}{\textbf{PerceptionDLM}}
& \textbf{DAM + COCONut + SAM} & \textbf{42.3} & \textbf{82.4} & \textbf{62.4} \\
\bottomrule
\end{tabular}
}
\end{table}

\noindent
\textbf{Ablations on Data Scaling}
~\Cref{tab:data_scaling_ablation} demonstrates consistent performance gains as the training data scales up. Starting with the DAM dataset baseline (53.7\% average accuracy), we observe that introducing processed data sourced from the COCONut dataset~\citep{deng2024coconut} improves the average accuracy to 57.7\%. Further expanding the training corpus with SA-1B(SAM) dataset~\citep{carion2025sam} annotations yields the best performance, reaching 62.4\%. Notably, the improvements indicate that both data scale and data diversity are critical.

\begin{table}[htbp]
\centering
\setlength{\belowcaptionskip}{8pt}
\caption{Ablations on the training strategy of the vision encoder in PerceptionDLM. Freezing the vision encoder performs better, and is thus used as the default baseline setting.}
%\vspace{2mm}
\label{tab:vision_encoder_ablation}
\setlength{\tabcolsep}{6pt}
\renewcommand{\arraystretch}{1.15}
\resizebox{0.75\textwidth}{!}{
\begin{tabular}{l | c | c c c}
\toprule
\textbf{Method} & \textbf{Vision Encoder} & \textbf{Pos(\%) $\uparrow$} & \textbf{Neg(\%) $\uparrow$} & \textbf{Avg(\%) $\uparrow$} \\
\midrule
& Train & 31.8 & 68.8 & 50.3 \\
\rowcolor{blue!8}
\cellcolor{white}\multirow{-2}{*}{\textbf{PerceptionDLM}}
& \textbf{Freeze} & \textbf{33.8} & \textbf{73.6} & \textbf{53.7} \\
\bottomrule
\end{tabular}
}
\end{table}

\noindent
\textbf{Ablations on Vision Encoder Training Strategy.}
We investigate whether unfreezing the vision encoder during visual instruction tuning benefits region-level perception. As shown in~\Cref{tab:vision_encoder_ablation}, fully updating the vision encoder results in performance degradation on the fine-grained ParaDLC-Bench, consistent with observations on multimodal benchmarks. This suggests that the robust representations learned during the vision encoder's large-scale pre-training are highly important for complex localized perception tasks.

% \begin{table}[htbp]
% \centering
% \setlength{\belowcaptionskip}{8pt}
% \caption{Ablation on the attention masking mechanism in PerceptionDLM. Structured attention yields better overall performance than full attention, and is therefore adopted as the default baseline setting.}
% %\vspace{2mm}
% \label{tab:attention_mask_ablation}
% \setlength{\tabcolsep}{6pt}
% \renewcommand{\arraystretch}{1.15}
% \resizebox{0.75\textwidth}{!}{
% \begin{tabular}{l | c | c c c}
% \toprule
% \textbf{Method} & \textbf{Attention Mask} & \textbf{Pos.(\%) $\uparrow$} & \textbf{Neg.(\%) $\uparrow$} & \textbf{Avg.(\%) $\uparrow$} \\
% \midrule
% % 第一行第一列留空
% & Full Attention & 30.8 & 64.1 & 47.5 \\
% \rowcolor{blue!8}
% % 使用白底覆盖蓝底，并使用向上跨行的 \multirow{-2} 渲染文字
% \cellcolor{white}\multirow{-2}{*}{\textbf{PerceptionDLM}} 
% & \textbf{Structured Attention} & 33.8 & \textbf{73.6} & \textbf{53.7} \\
% \bottomrule
% \end{tabular}
% }
% \end{table}

\begin{table}[htbp]
\centering
\setlength{\belowcaptionskip}{8pt}
\caption{Ablations on core design choices for parallel captioning under the same training setting, including module design and attention masking strategy. Results show that \textit{region prompting}, \textit{RoI-aligned feature replay}, and \textit{structured attention} are all important for the final performance. Following the judge-scoring protocol, Pos/Neg scores can be slightly negative when penalties dominate.}
%\vspace{2mm}
\label{tab:component_ablation}
\setlength{\tabcolsep}{6pt}
\renewcommand{\arraystretch}{1.15}
\resizebox{0.7\textwidth}{!}{
\begin{tabular}{l | c | c c c}
\toprule
\textbf{Method} & \textbf{Model Setting} & \textbf{Pos.(\%) $\uparrow$} & \textbf{Neg.(\%) $\uparrow$} & \textbf{Avg.(\%) $\uparrow$} \\
\midrule
& w/o region prompting & -0.01 & 2.4 & 1.1 \\
& w/o RoI-aligned feature replay & 29.9 & 72.7 & 51.3 \\
& Full Attention & 30.8 & 64.1 & 47.5 \\
\rowcolor{blue!8}
\cellcolor{white}\multirow{-4}{*}{\textbf{PerceptionDLM}}
& \textbf{Full model (default baseline)} & \textbf{33.8} & \textbf{73.6} & \textbf{53.7} \\
\bottomrule
\end{tabular}
}
\end{table}

\noindent
\textbf{Ablations on Architecture Designs.}
In~\Cref{tab:component_ablation}, we evaluate the necessity of our proposed structural designs for parallel captioning under a controlled training setting. The results confirm that all three core modules are indispensable:
\begin{itemize}
    \item Region Prompting: Without explicit region prompting, the model loses its spatial grounding capability entirely, resulting in a catastrophic drop to 1.1\% average accuracy. The model fails to associate textual generation with specific visual targets.
    \item Structured Attention: Replacing our structured attention with standard full attention degrades the average accuracy by 6.2\%. This validates that structured attention successfully isolates the processing of different masks during parallel generation, mitigating feature interference and target confusion across multiple concurrent descriptions.
    \item RoI-aligned Feature Replay: Removing this module results in a 2.4\% accuracy drop. This indicates that explicitly pooling and aligning localized visual features provides fine-grained cues that are essential for accurate regional perception.
\end{itemize}

\noindent
\textbf{Qualitative Breakdown of Region Prompting.}
Removing region prompting causes a catastrophic grounding failure, though not a language breakdown. Below is a qualitative example from the image in~\Cref{fig:teaser}(a) without learnable region prompts:
\begin{itemize}
    \item \textbf{Mask 1:} ``A tall, cylindrical stone pillar with a smooth surface and a slightly tapered top. The pillar has a uniform texture and appears to be made of a solid material.''
    \item \textbf{Mask 2:} ``A tall, cylindrical stone pillar with a smooth surface and a slightly tapered top...''
    \item \textbf{Mask 3:} ``A tall, cylindrical stone pillar with a smooth surface and a slightly tapered top...''
\end{itemize}
The model emits grammatically fluent, identical text for all masks. Without learnable region prompts, the model loses the ability to bind specific generated tokens to distinct spatial coordinates, completely failing the localized perception task.

\noindent
\textbf{Judge Model Sensitivity on ParaDLC-Bench.}
To establish reproducibility and test the robustness of our benchmark, we re-evaluate the models using different Judge LLMs, including the open-source Qwen3.5-27B and the closed-source Gemini-3.1-Pro. As shown in \Cref{tab:judge_sensitivity}, although there are minor fluctuations in absolute scores and partial rankings due to differences in the judge models' reasoning capabilities, the main conclusions remain unchanged. PerceptionDLM consistently outperforms all other diffusion Vision-Language Models, remaining highly competitive with traditional AR models. This fully validates the reliability of the ParaDLC-Bench evaluation framework.

\begin{table}[htbp]
\centering
\setlength{\belowcaptionskip}{8pt}
\caption{Performance on ParaDLC-Bench by different judge models (G3.1P = Gemini-3.1-Pro Judge; Q3.5 = Qwen3.5-27B Judge).}
\label{tab:judge_sensitivity}
\setlength{\tabcolsep}{4pt}
\renewcommand{\arraystretch}{1.15}
\resizebox{0.95\textwidth}{!}{
\begin{tabular}{l | c c c | c c c}
\toprule
\textbf{Model} & \textbf{G3.1P-Pos(\%)} & \textbf{G3.1P-Neg(\%)} & \textbf{G3.1P-Avg(\%)} & \textbf{Q3.5-Pos(\%)} & \textbf{Q3.5-Neg(\%)} & \textbf{Q3.5-Avg(\%)} \\
\midrule
GPT-5.2 & 42.6 & 72.6 & 57.6 & 50.7 & 72.8 & 61.8 \\
Gemini-2.5-Pro & 41.1 & 73.6 & 57.3 & 45.5 & 73.4 & 59.4 \\
Gemini-3.1-pro & 45.3 & 82.7 & 64.0 & 53.0 & 82.8 & 67.9 \\
PixelRefer & 46.9 & 87.7 & 67.3 & 57.3 & 88.8 & 73.1 \\
DAM & 52.3 & 88.0 & 70.1 & 59.6 & 88.8 & 74.2 \\
GAR & 53.5 & 88.8 & 71.2 & 59.7 & 89.7 & 74.7 \\
LLaDA-V & 31.0 & 53.0 & 42.0 & 38.9 & 51.0 & 45.0 \\
SDAR-VL & 37.8 & 32.7 & 35.3 & 44.4 & 38.1 & 41.3 \\
Dream-VL & 35.4 & 35.8 & 35.6 & 41.4 & 37.9 & 39.7 \\
\rowcolor{blue!8}
PerceptionDLM & 47.1 & 84.9 & 66.0 & 59.4 & 87.5 & 73.5 \\
\bottomrule
\end{tabular}
}
\end{table}

\noindent
\textbf{Impact of Denoising Steps on Accuracy and Latency.}
To illustrate the trade-off between accuracy and inference latency, we evaluate PerceptionDLM on ParaDLC-Bench using Qwen3.5-27B judge across varying diffusion steps while keeping the generation length fixed at 32. As shown in \Cref{tab:denoising_steps_ablation}, \texttt{steps=32} is the optimal operating point, offering the best balance of high captioning accuracy and efficient inference time.

\begin{table}[htbp]
\centering
\setlength{\belowcaptionskip}{8pt}
\caption{Impact of denoising steps on accuracy and latency on ParaDLC-Bench.}
\label{tab:denoising_steps_ablation}
\setlength{\tabcolsep}{6pt}
\renewcommand{\arraystretch}{1.15}
\resizebox{0.7\textwidth}{!}{
\begin{tabular}{l | c | c c c | c}
\toprule
\textbf{Method} & \textbf{Steps} & \textbf{Pos(\%) $\uparrow$} & \textbf{Neg(\%) $\uparrow$} & \textbf{Avg(\%) $\uparrow$} & \textbf{Time (s)} \\
\midrule
& 16 & 54.5 & 83.4 & 69.0 & 138 \\
\rowcolor{blue!8}
\cellcolor{white} & \textbf{32} & \textbf{59.4} & \textbf{87.5} & \textbf{73.5} & \textbf{276} \\
& 48 & 57.3 & 83.7 & 70.5 & 411 \\
\multirow{-4}{*}{\textbf{PerceptionDLM}} & 64 & 51.2 & 83.8 & 67.5 & 549 \\
\bottomrule
\end{tabular}
}
\end{table}

\noindent
\textbf{Inference with Exceeded Visual Prompts.}
To address how the model generalizes to mask counts exceeding the number of trained visual prompts, we conduct an additional experiment testing a forced edge case where the model reuses the exact same visual prompt for all masks during inference. As shown in \Cref{tab:exceeded_prompts}, forcing the model to reuse visual prompts for distinct targets results in a performance drop. However, this is a graceful degradation (the average accuracy evaluated by Qwen3.5-27B drops from 73.5\% to 68.6\%), indicating that the model retains a strong baseline of parallel perception even when prompt capacity is exceeded. 

For practical deployment in highly dense scenarios, the recommended strategy is to split the masks into multiple inference passes. As demonstrated by the parallelism scaling in~\Cref{fig:teaser}(c), this chunking strategy still yields a significant speedup over sequential AR pipelines while bypassing potential long-context limitations. Naturally, if a specific downstream application strictly requires simultaneous dense perception, the visual prompt capacity can easily be scaled up during training to accommodate it.  

\begin{table}[htbp]
\centering
\setlength{\belowcaptionskip}{8pt}
\caption{Inference with Exceeded Visual Prompts (Evaluated using Qwen3.5-27B Judge).}
\label{tab:exceeded_prompts}
\setlength{\tabcolsep}{6pt}
\renewcommand{\arraystretch}{1.15}
\resizebox{0.75\textwidth}{!}{
\begin{tabular}{l | c c c}
\toprule
\textbf{Model Setting} & \textbf{Q3.5-Pos(\%) $\uparrow$} & \textbf{Q3.5-Neg(\%) $\uparrow$} & \textbf{Q3.5-Avg(\%) $\uparrow$} \\
\midrule
PerceptionDLM (same visual prompt) & 52.8 & 84.4 & 68.6 \\
\rowcolor{blue!8}
\textbf{PerceptionDLM (baseline)} & \textbf{59.4} & \textbf{87.5} & \textbf{73.5} \\
\bottomrule
\end{tabular}
}
\end{table}

\noindent
\textbf{Ablations on Caption Length Scaling.}
To investigate the impact of description length, we conduct an ablation study by scaling the maximum number of generation tokens per mask from 32 to 64, while keeping the diffusion steps fixed at 32. As shown in \Cref{tab:length_scaling}, with fixed denoising steps, generating longer sequences forces the model to decode more tokens per step. This increased decoding density leads to severe error accumulation along the denoising trajectory, resulting in semantic drifting.

\begin{table}[htbp]
\centering
\setlength{\belowcaptionskip}{8pt}
\caption{Caption length scaling on Parallel Captioning(Evaluated using Qwen3.5-27B Judge).}
\label{tab:length_scaling}
\setlength{\tabcolsep}{6pt}
\renewcommand{\arraystretch}{1.15}
\resizebox{0.75\textwidth}{!}{
\begin{tabular}{l | c | c | c c c}
\toprule
\textbf{Method} & \textbf{Length} & \textbf{Q3.5-Pos(\%) $\uparrow$} & \textbf{Q3.5-Neg(\%) $\uparrow$} & \textbf{Q3.5-Avg(\%) $\uparrow$} \\
\midrule
& 64 & 55.5 & 80.4 & 68.0 \\
\rowcolor{blue!8}
\cellcolor{white}\multirow{-2}{*}{\textbf{PerceptionDLM}} & \textbf{32} & \textbf{59.4} & \textbf{87.5} & \textbf{73.5} \\
\bottomrule
\end{tabular}
}
\end{table}

\noindent
\textbf{Single-Region Quality Regression Ablation.}
To investigate whether parallel training degrades single-region capabilities, we conduct an ablation study training PerceptionDLM exclusively on single-mask format data. As shown in \Cref{tab:single_mask_ablation}, single-mask training yields performance similar to our parallel training baseline. This indicates that the gap compared to AR models on single-region tasks is not a regression caused by parallelization.

\begin{table}[htbp]
\centering
\setlength{\belowcaptionskip}{8pt}
\caption{Single Mask Training Ablation on DLC-Bench.}
\label{tab:single_mask_ablation}
\setlength{\tabcolsep}{6pt}
\renewcommand{\arraystretch}{1.15}
\resizebox{0.75\textwidth}{!}{
\begin{tabular}{l | c c c}
\toprule
\textbf{Method} & \textbf{Pos(\%) $\uparrow$} & \textbf{Neg(\%) $\uparrow$} & \textbf{Avg(\%) $\uparrow$} \\
\midrule
PerceptionDLM (single mask training) & \textbf{34.5} & 69.8 & 52.1 \\
\rowcolor{blue!8}
\textbf{PerceptionDLM (baseline)} & 33.4 & \textbf{72.8} & \textbf{53.1} \\
\bottomrule
\end{tabular}
}
\end{table}

\noindent
\textbf{Zero-Shot Performance of PerceptionDLM-Base on ParaDLC-Bench.}
To disentangle the benefits of our proposed architectural design from the supervisory signals provided by GAR-8B in the ParaCaption-5.7M dataset, we evaluate the zero-shot region perception capability of PerceptionDLM-Base prior to any dataset-specific fine-tuning. Evaluated by the GPT-5.2 judge, PerceptionDLM-Base achieves a zero-shot average accuracy of \textbf{53.0\%} on ParaDLC-Bench. This baseline substantially outperforms existing DVLMs, such as SDAR-VL, which attains an average accuracy of 31.3\%(shown in~\Cref{tab:dlcbench_comparison}). This demonstrates that the significant performance improvements of our framework are fundamentally driven by the architectural advancements, rather than merely stemming from the high-quality synthetic training corpus.

\section{Details of ParaDLC-Bench} \label{sec:detail_bench}
\subsection{Image and Instance Selection}
To ensure the diversity and complexity of the evaluation scenarios, our base images are sourced from two mainstream open-source datasets: the validation set of Objects365 V2~\citep{shao2019objects365} and the DaTaSeg Objects365 Instance Segmentation Dataset~\citep{gu2023dataseg}. Both datasets provide a large number of images with high-quality instance segmentation masks. Specifically, the Objects365 V2 subset includes 54 images with a total of 178 mask instances, while the DaTaSeg subset comprises 46 images with a total of 121 mask instances.

Regarding the instance sampling and filtering criteria, we primarily rely on the professional judgment of human annotators. Unlike DLC-Bench ~\citep{lian2025describe}, which focuses on a single challenging target, ParaDLC-Bench specifically emphasizes the interaction potential among multiple targets. Annotators consciously select "multi-instance combinations" that are spatially adjacent, semantically confusing, or prone to feature entanglement. Finally, to ensure the benchmark's rigor, we performed strict data deduplication and isolation, ensuring that none of the images used in this benchmark were present in the training sets of the evaluated models, thereby preventing data leakage.

\subsection{Data Annotation Pipeline}
\begin{figure*}[htbp]
    \centering
    \includegraphics[width=1.\linewidth]{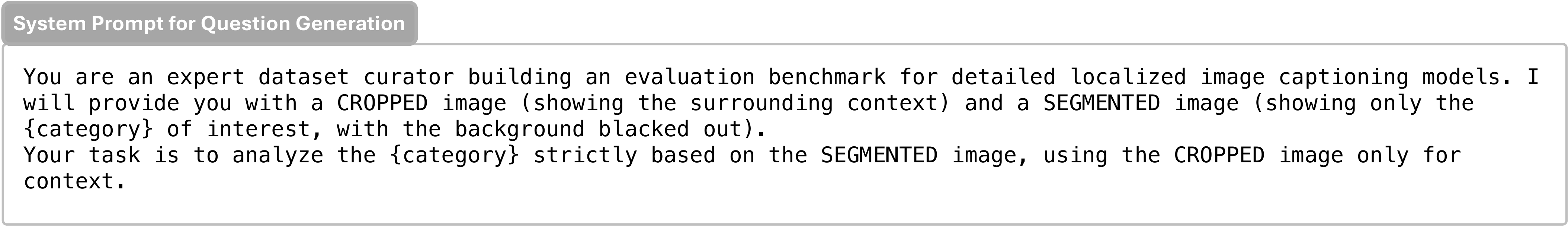}
    \caption{System Prompt for Question Generation}
    %\vspace{-2mm}
    \label{fig:prompt1}
\end{figure*}

\begin{figure*}[htbp]
    \centering
    \includegraphics[width=1.\linewidth]{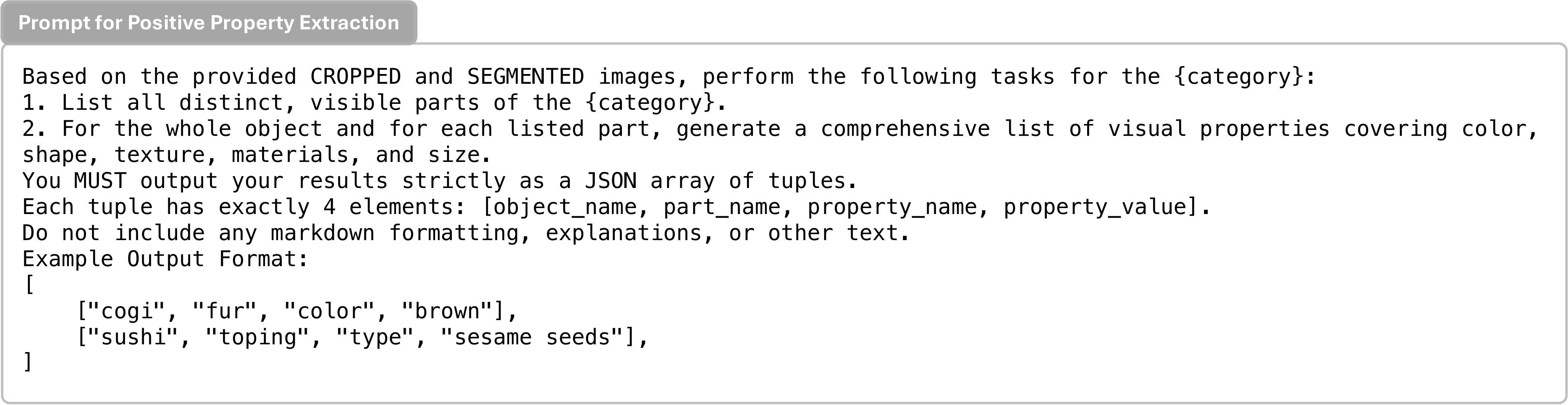}
    \caption{Prompt for Positive Property Extraction}
    %\vspace{-2mm}
    \label{fig:prompt2}
\end{figure*}

\begin{figure*}[htbp]
    \centering
    \includegraphics[width=1.\linewidth]{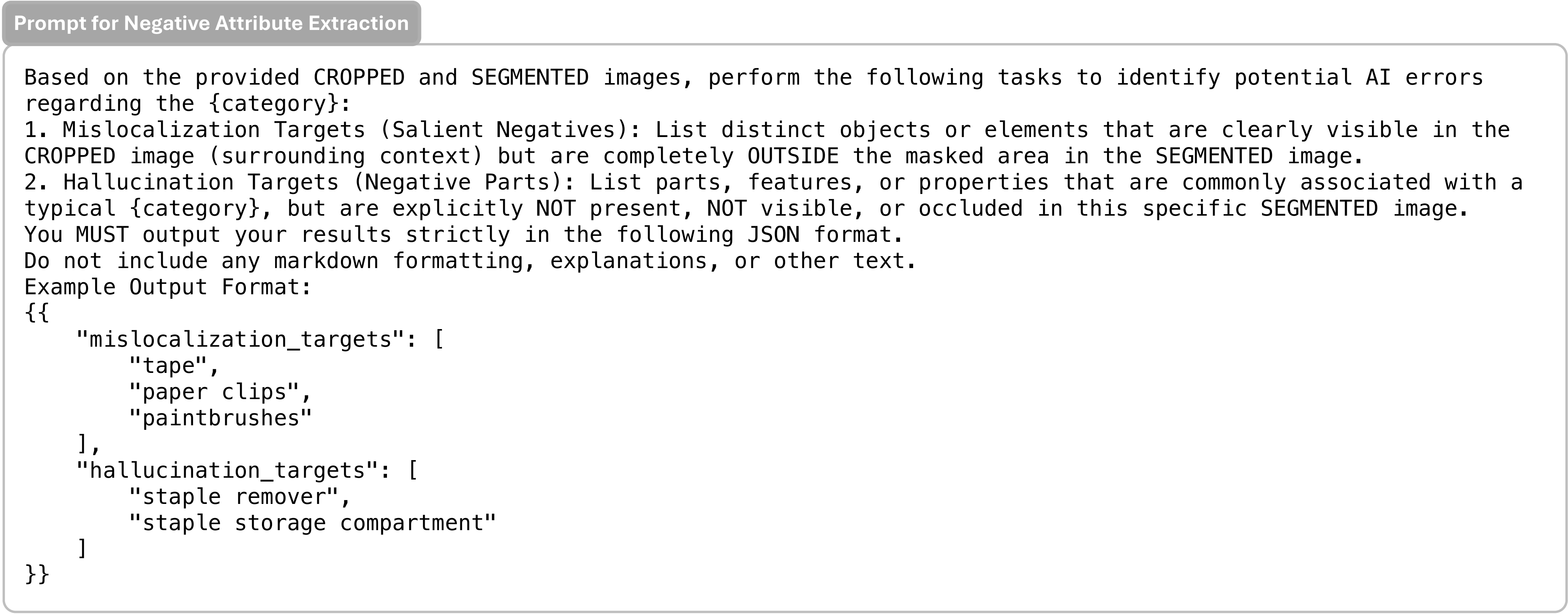}
    \caption{Prompt for Negative Attribute Extraction}
    %\vspace{-2mm}
    \label{fig:prompt3}
\end{figure*}

\begin{figure*}[htbp]
    \centering
    \includegraphics[width=1.\linewidth]{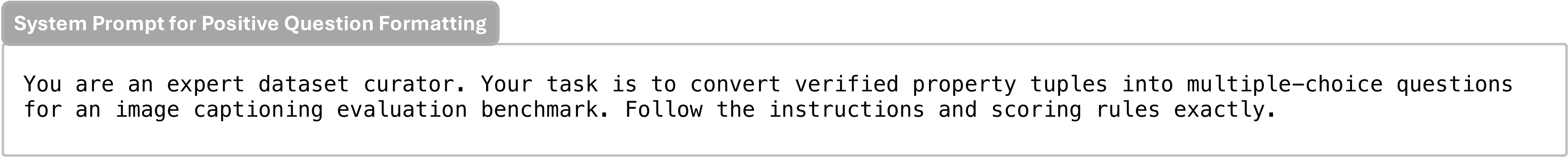}
    \caption{System Prompt for Positive Question Formatting}
    %\vspace{-2mm}
    \label{fig:prompt4}
\end{figure*}

\begin{figure*}[htbp]
    \centering
    \includegraphics[width=1.\linewidth]{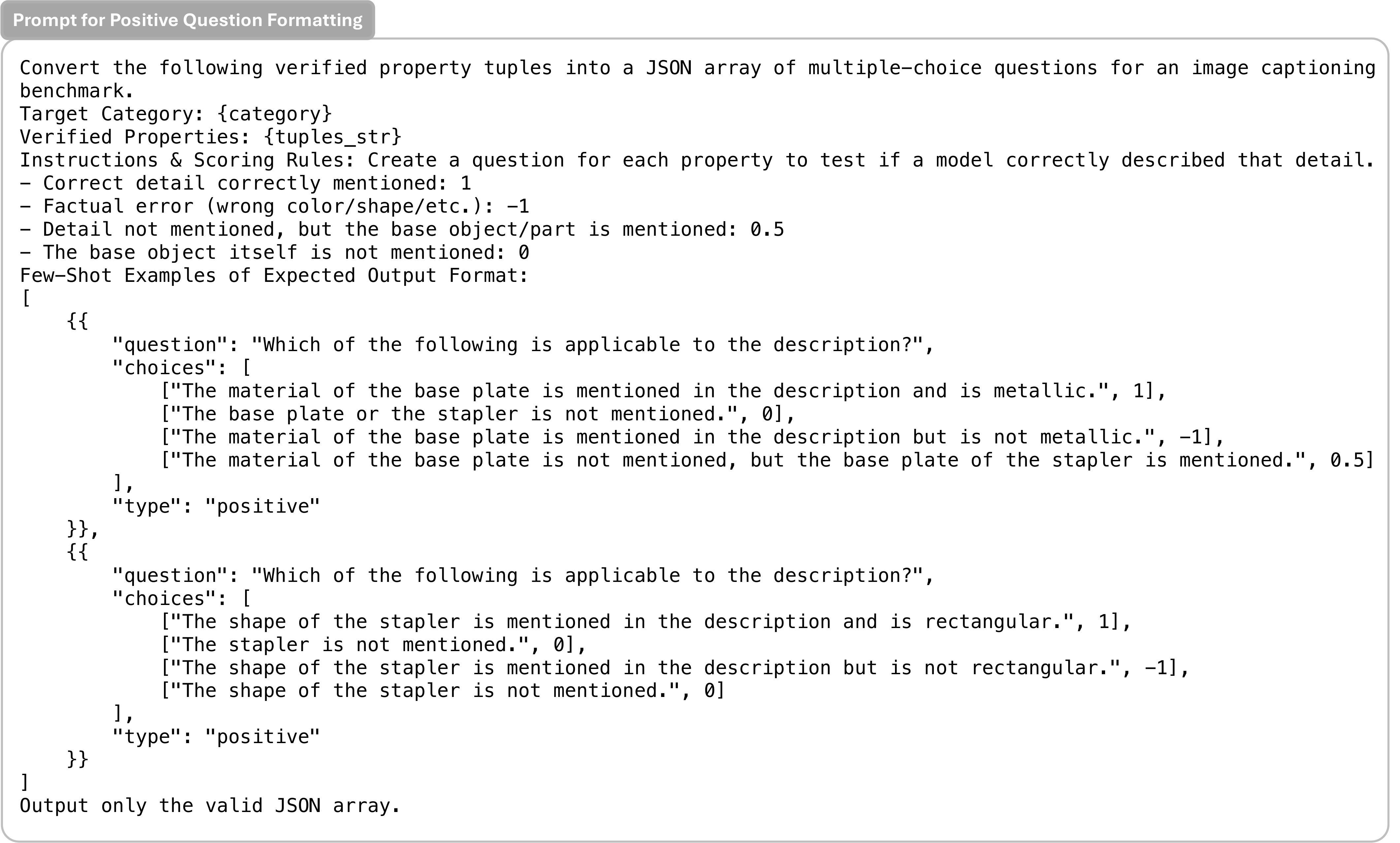}
    \caption{Prompt for Positive Question Formatting}
    %\vspace{-2mm}
    \label{fig:prompt5}
\end{figure*}

\begin{figure*}[htbp]
    \centering
    \includegraphics[width=1.\linewidth]{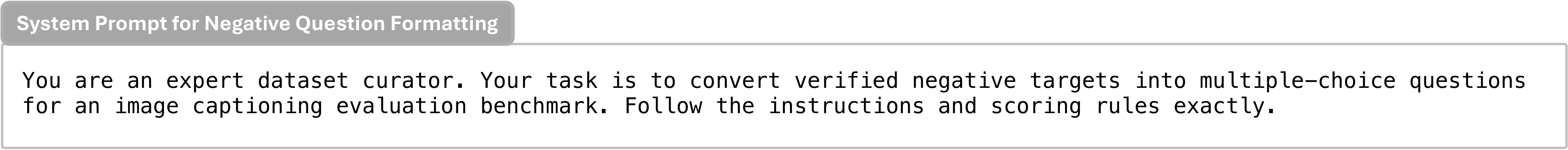}
    \caption{System Prompt for Negative Question Formatting}
    %\vspace{-2mm}
    \label{fig:prompt6}
\end{figure*}

\begin{figure*}[htbp]
    \centering
    \includegraphics[width=1.\linewidth]{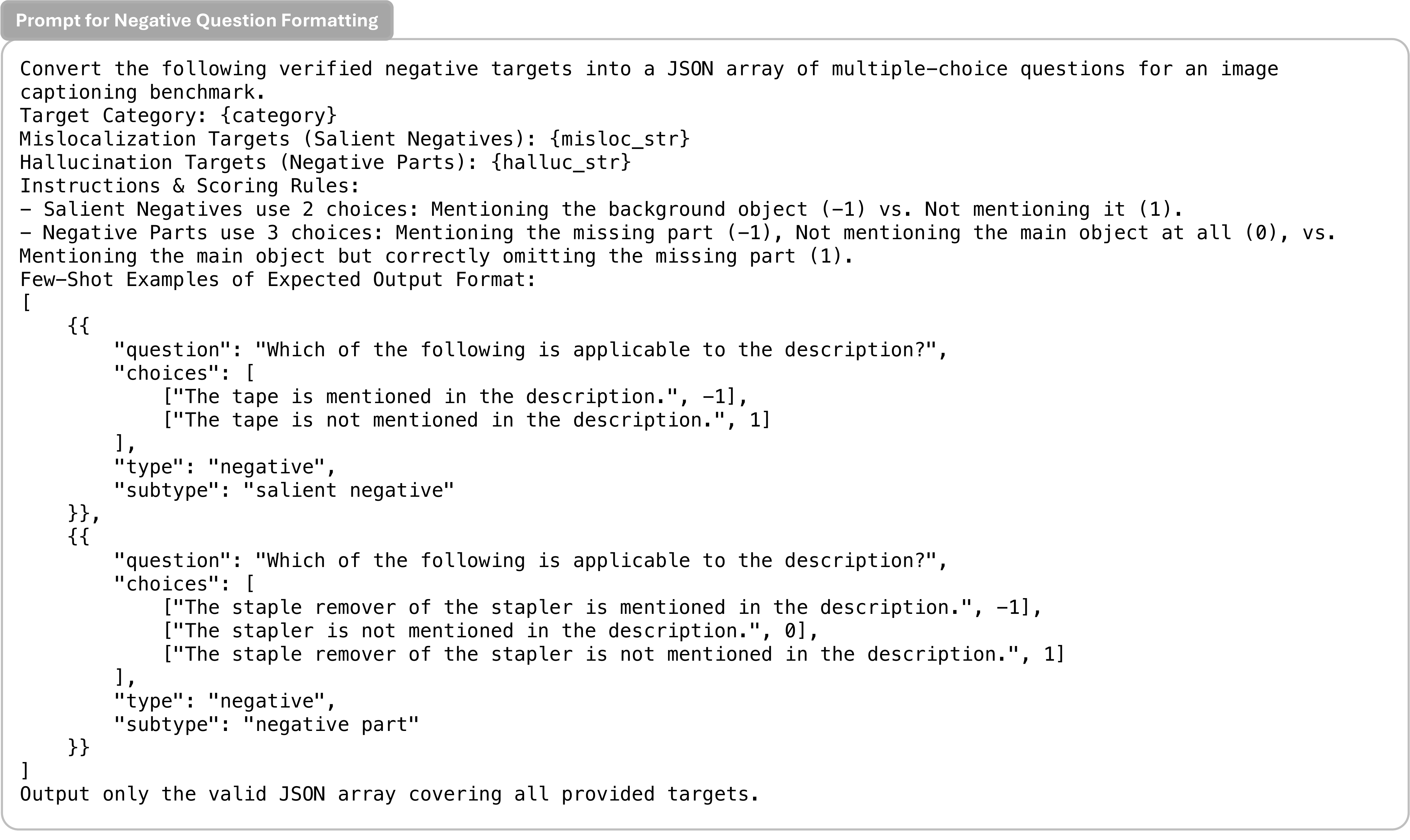}
    \caption{Prompt for Negative Question Formatting}
    %\vspace{-2mm}
    \label{fig:prompt7}
\end{figure*}

The question generation pipeline of ParaDLC-Bench is inspired by the methodology of DLC-Bench~\citep{lian2025describe}, where the goals are designed specifically for "multi-target scenarios". The entire process consists of three core components: "LLM Extraction - Human Curation - Automated Format Conversion". 
We uniformly use the more advanced GPT-5.2~\citep{gpt5.2} as the foundational model for complex logical reasoning. 
We provide all of our prompts utilized in constructing our ParaDLC-Bench in~\Cref{fig:prompt1,fig:prompt2,fig:prompt3,fig:prompt4,fig:prompt5,fig:prompt6,fig:prompt7}.

\textbf{Attribute Extraction:} We input the cropped version of the original image alongside the segmented image (retaining only the target instance) into GPT-5.2. For positive attribute extraction, the model is required to extract all visible parts of the target and generate an attribute list covering dimensions such as color, shape, texture, material, and size, uniformly outputting a tuple in the format of \texttt{([object name], [part name], [property name], [property value])}. Regarding negative attribute extraction—our core innovative design—we construct challenging distractors tailored for multi-target scenarios by extracting two types of negative samples that easily trigger model hallucinations: first, \textit{Negative Parts}, which are parts commonly associated with the object but invisible in the current mask; second, \textit{Salient Negatives / Mislocalization Targets}. In ParaDLC-Bench, the latter not only includes standard background objects but also focuses heavily on other concurrently evaluated targets in the image. This aims to test whether the model will incorrectly assign the features of other concurrent targets to the current one.

\begin{figure*}[!t]
    \centering
    \includegraphics[width=1.\linewidth]{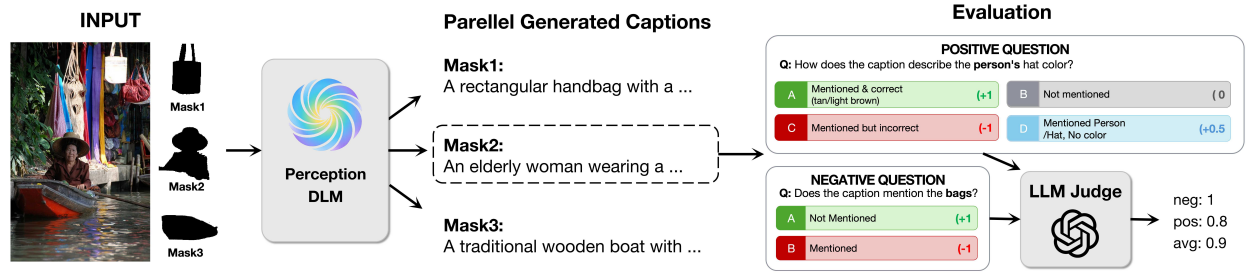}
    \caption{\textbf{The evaluation pipeline for detailed localized captioning in ParaDLC-Bench.} (a) During evaluation, given an input image and multiple region masks, the Perception DLM simultaneously generates textual descriptions for all specified regions in a single pass. (b) Subsequently, each generated caption, paired with its corresponding evaluation questions, is fed into a text-only large language model (LLM Judge) to calculate the final score. (c) In the scoring mechanism, positive questions reward accurate details and penalize factual errors, while negative questions strictly penalize mislocalization and hallucinations to prevent false positives in irrelevant areas.}
    \label{fig:bench}
\end{figure*}

\textbf{Human Curation \& Refinement:} To compensate for factual errors made by the LLM, human experts perform strict cross-verification starting from the initial lists generated by GPT-5.2. Experts are responsible for supplementing missing salient attributes and removing ambiguous details, ensuring that all attributes strictly align with the masked region and that the distractors are highly deceptive.

\textbf{Question Generation:} After obtaining the manually verified attribute lists, we utilize GPT-5.2 to convert them into mutually exclusive multiple-choice questions. Through this rigorous pipeline, we ultimately constructed an evaluation question bank comprising 2,345 high-quality multiple-choice questions.

\subsection{Scoring Mechanism}
During the evaluation phase, we inherit the reference-free judge evaluation paradigm from DLC-Bench~\citep{lian2025describe}. The evaluated models are required to generate descriptions for multiple specified masked regions in the image. For models equipped with parallel generation capabilities, descriptions for all regions can be generated concurrently in a single pass; for models lacking this feature, descriptions are generated sequentially for each masked region. 
Subsequently, an LLM Judger (GPT-5.2~\citep{gpt5.2}) evaluates these descriptions using predefined positive and negative questions. 
Regarding judge variance, GPT-5.2 evaluations are conducted at strict \texttt{temperature=0} to eliminate sampling noise.

To prevent models from cheating to gain higher scores by generating excessively long and ambiguous descriptions, a model can only receive positive or negative points if the base target is correctly identified. Unlike DLC-Bench, which averages global positive and negative scores, we compute the final score by averaging the individual scores across all masks, ensuring equal evaluation weighting for each region. The overall scoring process and an example are illustrated in~\Cref{fig:bench}.

\begin{figure*}[htbp]
    \centering
    \includegraphics[width=0.8\linewidth]{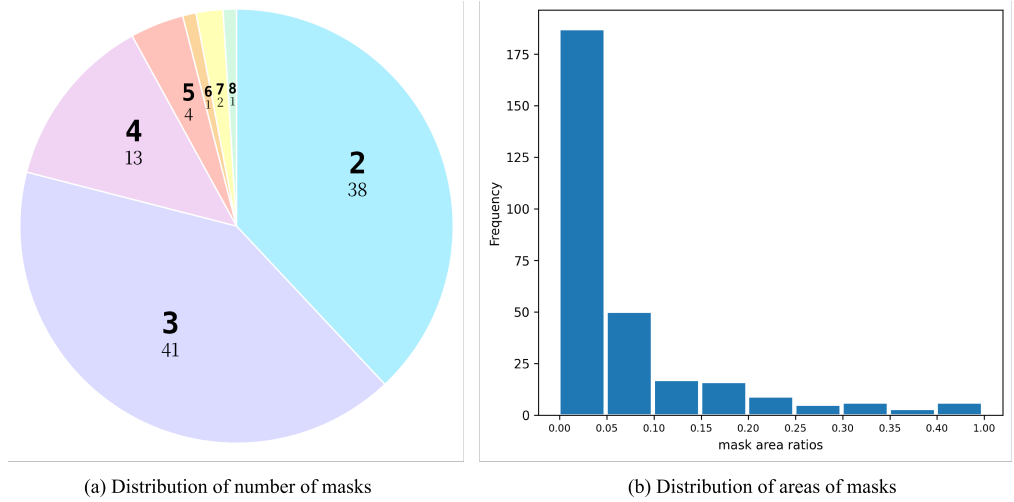}
    \caption{Data statistics of ParaDLC-Bench}
    %\vspace{-2mm}
    \label{fig:statics}
\end{figure*}

\subsection{Data Statistics}
To intuitively illustrate the data composition and task difficulty of ParaDLC-Bench, we conducted a multi-dimensional statistical analysis on the finalized dataset:

\textbf{Distribution of Number of Masks:} To highlight the core setting of "multi-target interaction,"~\Cref{fig:statics} details the distribution of the number of tested masks contained within a single image. All images in this benchmark contain 2 or more masked regions, with the vast majority ranging from 2 to 4. Some extremely challenging samples even contain up to 8 masks. This high-density distribution severely tests the model's local perception and anti-interference capabilities within complex contexts.

\textbf{Distribution of Areas of Masks:}~\Cref{fig:statics} presents the area proportion distribution of the masked regions relative to the entire image. 
To fully test the model's local description capabilities at a fine-grained level, we intentionally retained a large number of micro-detail targets (small masks) that occupy an extremely small proportion of the image. 
The average mask area ratio is 0.07. This distribution highlights the importance of addressing small-scale and fine-grained understanding.

\section{Related Work}
\label{sec:related_work}
% 1. Diffusion Language Model.
% 2. Image caption, ar->dvlm, parallel
% 3. Region understanding.

\noindent
\textbf{Diffusion Language Models.} Diffusion Language Models (DLMs) have recently emerged as a promising alternative to autoregressive language modeling. 
Among them, masked diffusion models~\citep{you2025llada, ye2025dream, yangmmada}, have shown particularly strong empirical performance.
Building on this line of work, LLaDA scales masked diffusion language modeling to 8B parameters and demonstrates that diffusion-based language models can approach the performance of strong autoregressive LLMs, such as LLaMA3-8B, across a wide range of downstream tasks.
Recent LLaDA series further scale the model size to over 100B parameters, with a more complex architecture (Mixture of Experts, MoE) and stronger reasoning capabilities.
Meanwhile, several works also extend DLMs to multimodal settings, including visual understanding and generation~\citep{yangmmada, xin2025lumina}.
LLaDA-V~\citep{you2025llada} adapts the visual instruction tuning paradigm to masked diffusion language modeling, while MMaDA~\citep{yangmmada} extends DLMs to both generation and understanding.
After that, several works explore AR-based adaptation by leveraging pre-trained AR knowledge~\citep{cheng2025sdar, bie2025llada2}.
Our work mainly focuses on multimodal understanding tasks, where we build a stronger baseline that sets a new state of the art among diffusion VLMs.

\noindent
\textbf{Image Captioning in MLLMs.} Most image caption datasets are used for multimodal pre-training tasks, which is the first stage of MLLMs for text and vision alignment~\citep{li2025llava}.
With the rapid development of MLLMs, this is one of their abilities.
Recent works explore various caption settings, including dense, detailed, grounded, and region-sensitive formats~\citep{li2025denseworld, wang2025grasp}.
All previous solutions focus on AR-based image-to-text generation.
Several multimodal DLMs can generate text in parallel.
In contrast, our goal is to explore multiple-object caption generation in parallel using DLMs, enabling parallel generation at both the sequence (object) and token levels.
To our knowledge, we are the first to achieve this function.

\noindent
\textbf{Region Understanding of MLLMs.}
In contrast to conventional MLLMs that primarily emphasize image-level understanding without dedicated extraction of localized features, region-level understanding requires MLLMs to accurately model region-specific attributes, fine-grained visual details, and spatial relationships within designated areas. 
Consequently, enhancing regional perception has become a central research focus for advancing MLLMs' visual perception capabilities.
Existing works have adopted three mainstream strategies to represent image regions-of-interest (RoIs), namely visual markers~\citep{yang2023set}, bounding boxes~\citep{zhang2307gpt4roi, chen2023shikra, rasheed2024glamm}, and segmentation masks~\citep{zhang2024omg,yuan2025sa2va, zhang2025pixel,yuan2025visual,wang2025grasp, zhou2026samtok, lian2025describe}.
These models are predominantly designed for single-region, single-caption generation, and nearly all adopt autoregressive (AR) LLMs as their core inference backbone, which restricts concurrent multi-region processing.
In contrast, our approach leverages DLMs that offer inherent efficiency advantages, thereby enabling natural support for high-efficiency \textit{parallel} region caption generation.
\label{sec:related_works}

\section{Visualization and Qualitative Results}
\label{sec:qua}

\noindent
\textbf{Parallel Perception Capabilities.} 
In~\Cref{fig:cases}, we present qualitative examples demonstrating that PerceptionDLM can simultaneously generate highly detailed and accurate descriptions for multiple masked regions within a single image through a single forward inference pass. Whether dealing with tightly packed objects (e.g., the various ingredients in the bowl) or spatially distributed subjects (e.g., the man and his fishing rod), the model successfully captures fine-grained attributes such as color, texture, shape.

\begin{figure*}[htbp]
    \centering
    \includegraphics[width=1.\linewidth]{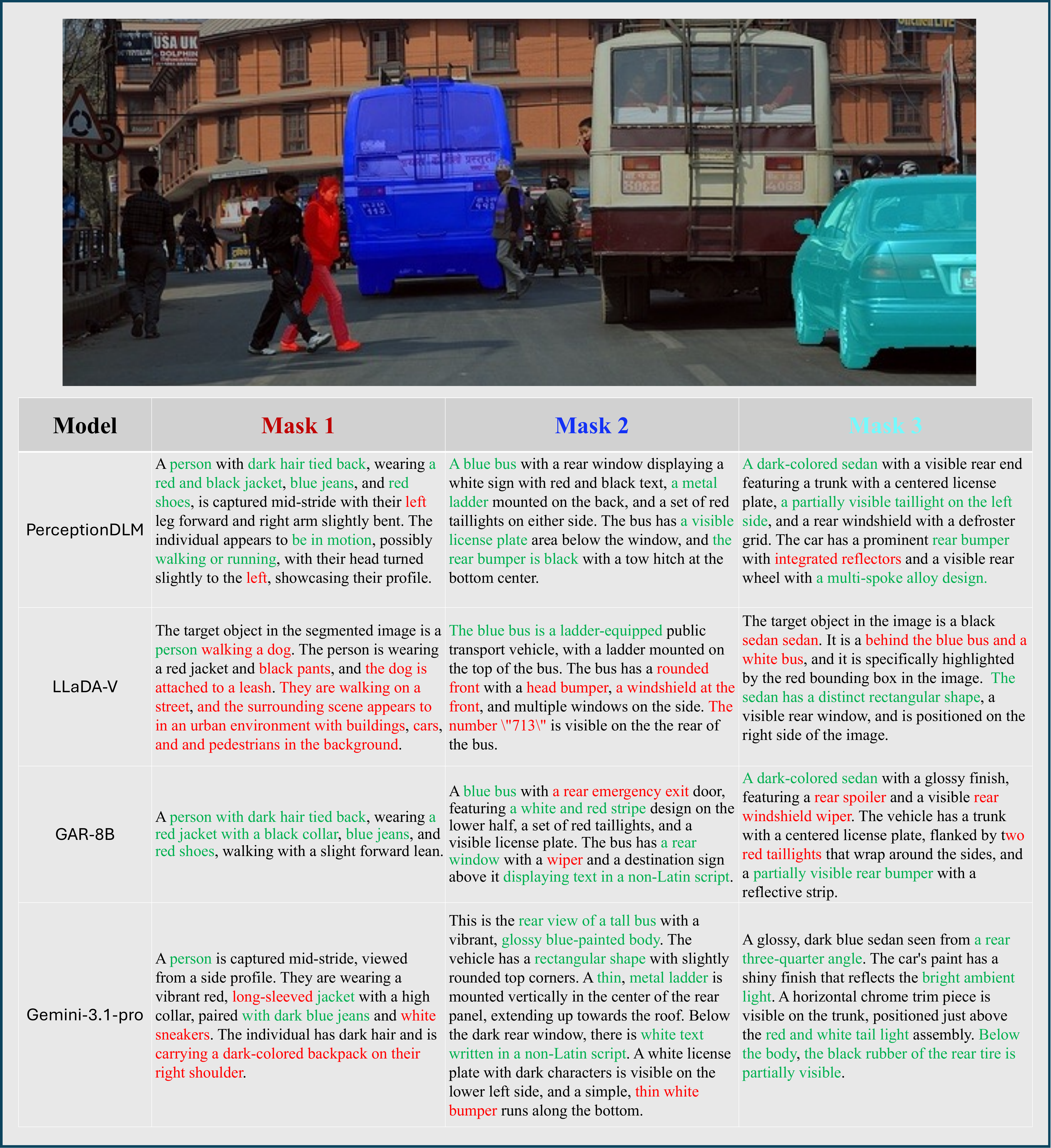}
    \caption{\textbf{Qualitative comparison of PerceptionDLM against existing baselines.} Correct attribute descriptions are highlighted in green, whereas hallucinations and errors are marked in red. PerceptionDLM effectively mitigates the severe cross-region interference observed in other models.}
    %\vspace{-2mm}
    \label{fig:compare}
\end{figure*}

\noindent
\textbf{Comparison with Baseline Models.} 
In~\Cref{fig:compare}, we compare PerceptionDLM against existing strong baselines, including LLaDA-V, GAR-8B, and Gemini-3.1-Pro. When faced with adjacent and visually complex masks, PerceptionDLM effectively isolates region-specific visual context, producing descriptions that strictly align with the intended target regions. In contrast, baseline models frequently suffer from severe hallucinations and feature confusion. For instance, standard diffusion VLMs and AR-based region models may mistakenly assign the attributes of a neighboring object (e.g., colors, text, or parts) to the current target mask, or hallucinate entirely non-existent background elements. By comparison, PerceptionDLM significantly reduces such errors, effectively mitigating the severe cross-region interference observed in other models.

\begin{figure*}[htbp]
    \centering
    \includegraphics[width=0.8\linewidth]{figs/infer_case.pdf}
    \caption{\textbf{Qualitative examples} of PerceptionDLM generating detailed descriptions for multiple masked regions in parallel.}
    %\vspace{-2mm}
    \label{fig:cases}
\end{figure*}

\section{Limitations and Future Works}
\label{sec:limitation}
While PerceptionDLM significantly improves multi-region perception efficiency, certain limitations remain. First, the actual inference speed is still bounded by the multi-step denoising process inherent to diffusion models. Future work could address this by exploring step distillation techniques (e.g., pseudo-trajectory distillation~\citep{qian2026d3llm}) to compress the generation into fewer steps. Second, eliminating attribute entanglement in extremely dense or semantically similar regions remains challenging; introducing explicit region contrastive losses or more fine-grained attention masking could further mitigate hallucinations. Finally, to bridge the performance gap in complex reasoning tasks compared to advanced autoregressive models, applying Reinforcement Learning (RL) techniques to unlock deep reasoning capabilities represents a critical direction for future exploration.

\begin{figure*}[htbp]
    \centering
    \begin{minipage}{0.49\linewidth}
        \centering
        \includegraphics[width=\linewidth]{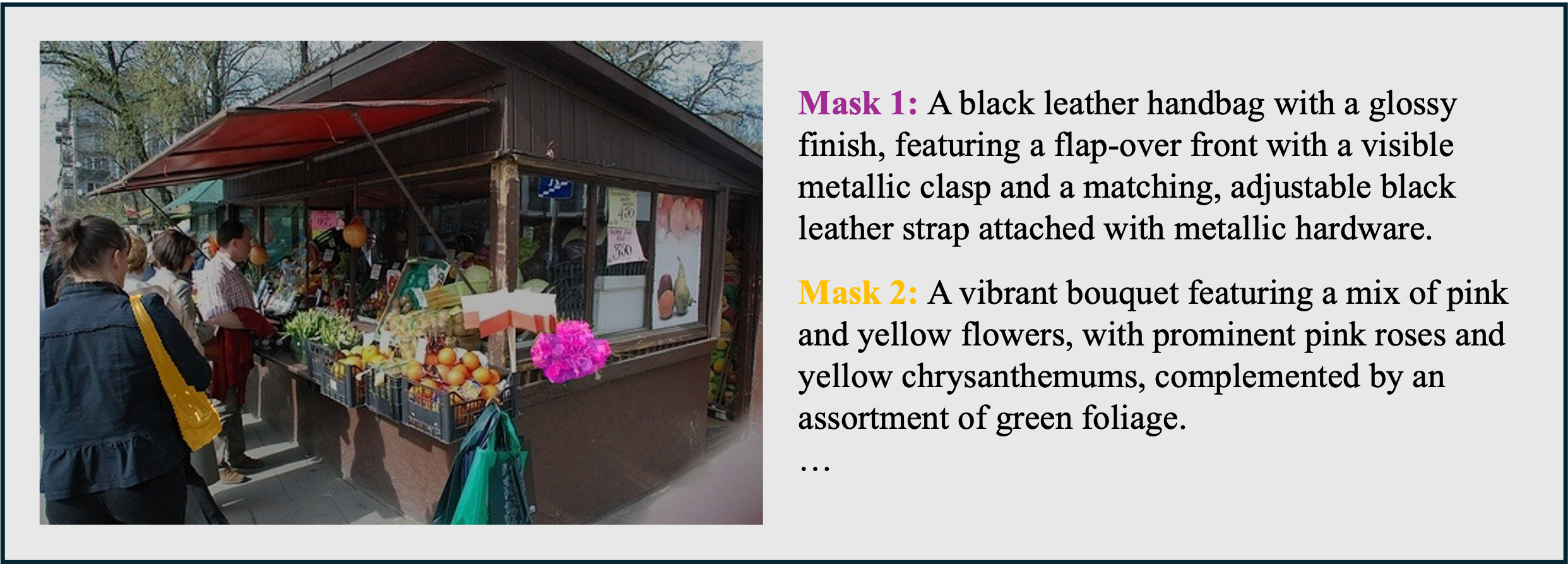}
        \vspace{1mm}
        \centerline{(a) Cross-region attribute entanglement}
    \end{minipage}
    \hfill
    \begin{minipage}{0.49\linewidth}
        \centering
        \includegraphics[width=\linewidth]{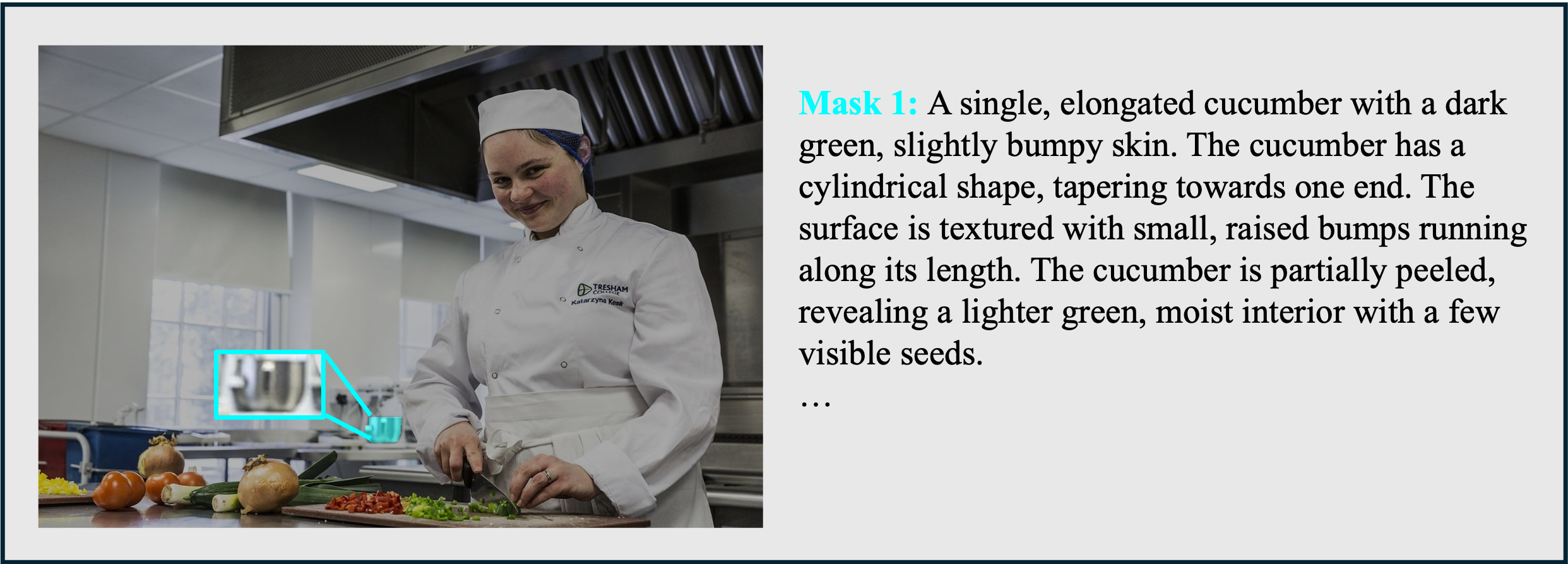}
        \vspace{1mm}
        \centerline{(b) Tiny or heavily occluded regions}
    \end{minipage}

    \vspace{3mm}

    \begin{minipage}{0.49\linewidth}
        \centering
        \includegraphics[width=\linewidth]{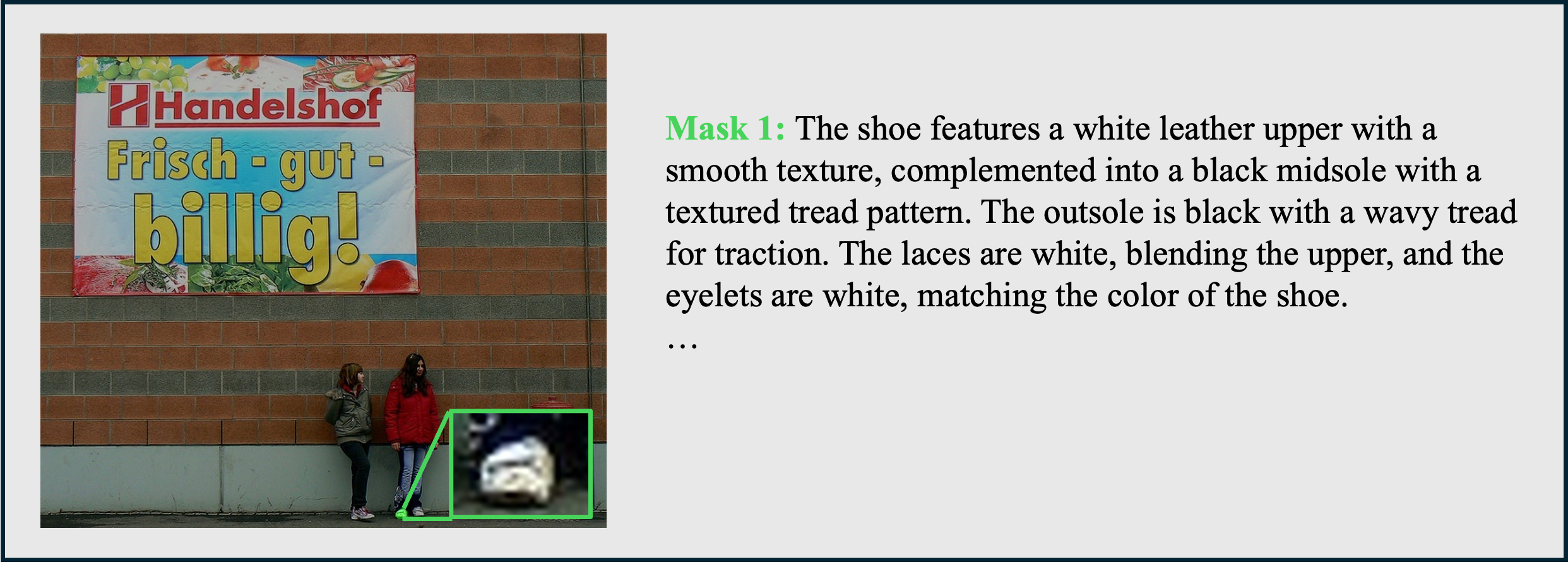}
        \vspace{1mm}
        \centerline{(c) Hallucination of typical-but-absent attributes}
    \end{minipage}
    \hfill
    \begin{minipage}{0.49\linewidth}
        \centering
        \includegraphics[width=\linewidth]{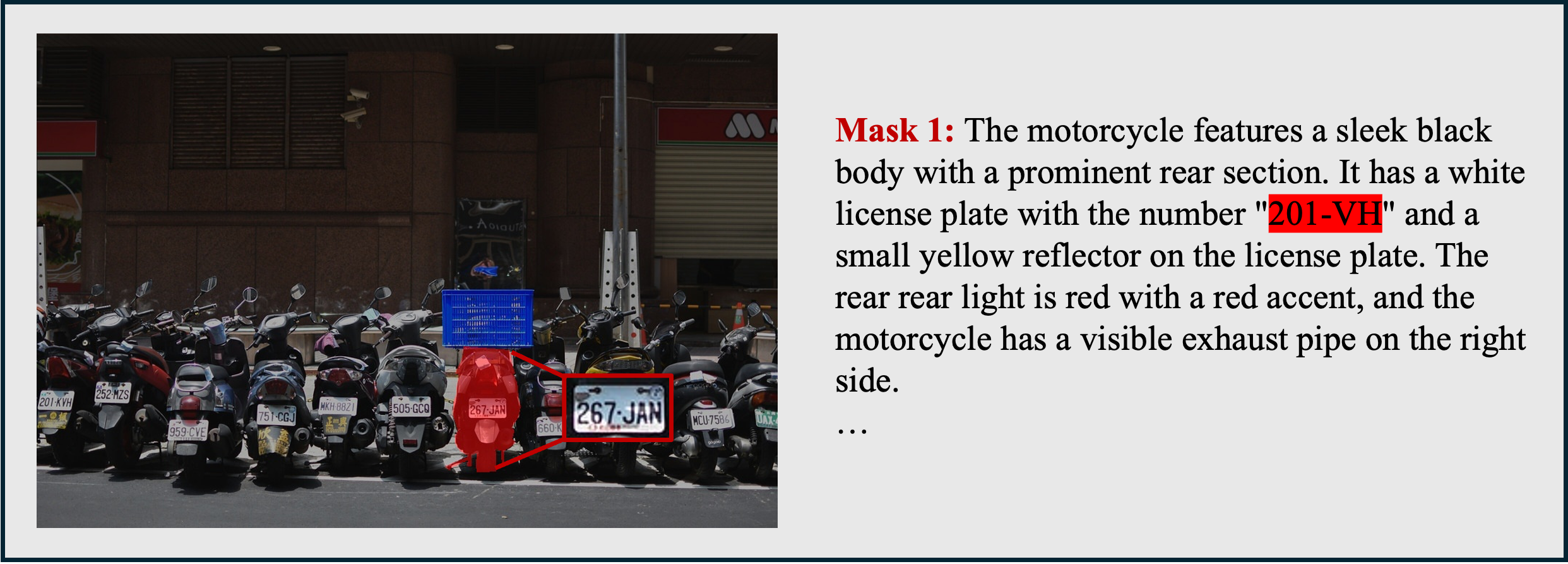}
        \vspace{1mm}
        \centerline{(d) Fine-grained text}
    \end{minipage}

    \caption{Representative failure cases of PerceptionDLM.}
    \label{fig:failure_cases}
\end{figure*}

\noindent
\textbf{Failure Case Analysis.}
To better characterize the current limitations, we conduct a qualitative analysis of typical failure cases and group them into four representative categories, as illustrated in~\Cref{fig:failure_cases}:
\begin{itemize}
    \item \textbf{(a) Cross-region attribute entanglement.} When multiple masks are spatially adjacent or semantically similar, the model occasionally leaks attributes (e.g., color, texture, or material) from a neighboring target into the current region's description. For example, in~\Cref{fig:failure_cases}(a), the caption of one mask absorbs the appearance of an adjacent object on the same market stall.
    \item \textbf{(b) Tiny or heavily occluded regions.} For extremely small or heavily occluded masks that occupy a negligible fraction of the image, limited visual evidence can lead to vague or partially incorrect descriptions. For example, in~\Cref{fig:failure_cases}(b), a small, low-resolution region is given an overly specific yet inaccurate description.
    \item \textbf{(c) Hallucination of typical-but-absent attributes.} Following the two negative types in ParaDLC-Bench, the model may fabricate parts commonly associated with the object category yet absent from the mask, or borrow attributes from co-occurring objects. For example, in~\Cref{fig:failure_cases}(c), it invents white laces, eyelets, and tread for a tiny, blurry shoe region that are indiscernible from the pixels.
    \item \textbf{(d) Fine-grained text.} The model can still struggle with reading small embedded text (OCR). For example, in~\Cref{fig:failure_cases}(d), it misreads the license plate as ``201-VH'' instead of the actual ``267-JAN''.
\end{itemize}
These observations are consistent with our quantitative findings and motivate the future directions discussed above.

% \section{Prompts Templates}
% \label{sec:prompt}
% We provide all of our prompts utilized in building our data in \Cref{}

% \section{Use of LLMs}
% \label{sec:llm}

% In preparing this paper, LLMs are employed in the methodology and preparation of this research. Specifically, LLMs are integrated into the automated pipelines to facilitate data construction, attribute extraction, and evaluation for our proposed datasets and benchmarks. During the drafting of this manuscript, AI-assisted tools are utilized strictly for lightweight tasks such as proofreading, grammatical correction, language refinement, and code implementation. All core research ideas, experimental designs, and substantive scientific writing remain the authors' original work.

\end{document}